\documentclass{article}

\usepackage{arxiv}

\usepackage[utf8]{inputenc} % allow utf-8 input
\usepackage[T1]{fontenc}    % use 8-bit T1 fonts
\usepackage[hidelinks]{hyperref}
\usepackage{url}            % simple URL typesetting
\usepackage{booktabs}       % professional-quality tables
\usepackage{amsfonts}       % blackboard math symbols
\usepackage{nicefrac}       % compact symbols for 1/2, etc.
\usepackage{microtype}      % microtypography
\usepackage{cleveref}       % smart cross-referencing
\usepackage{lipsum}         % Can be removed after putting your text content
\usepackage{graphicx}
\usepackage{natbib}
\usepackage{doi}
\usepackage{lineno}         % Added for line numbers
\usepackage{longtable}
\usepackage{array}
\usepackage{makecell}
\usepackage{xcolor}
\usepackage{pifont}          
\usepackage{fontawesome5}    
\usepackage{ragged2e} 

\newcommand{\cmark}{\textcolor{green!55!black}{\ding{51}}}
\newcommand{\xmark}{\textcolor{red!70!black}{\ding{55}}}

\newcommand{\iconOpenWater}{\textcolor{blue!60!black}{\small\faWater}}    % wave
\newcommand{\iconVesselSite}{\textcolor{teal!70!black}{\small\faShip}}    % ship
\newcommand{\iconTurbine}{\textcolor{green!55!black}{\small\faFan}}       % fan = turbine rotor
\newcommand{\iconPlatform}{\textcolor{orange!80!black}{\small\faIndustry}}% industrial structure
\newcommand{\iconFoundation}{\textcolor{brown!70!black}{\small\faHockeyPuck}}  % hockey puck = foundation
\newcommand{\iconMooring}{\textcolor{purple!70!black}{\small\faAnchor}}   % anchor = mooring

\title{Global Offshore Wind Infrastructure: Deployment and Operational Dynamics from Dense Sentinel-1 Time Series}

\newif\ifuniqueAffiliation
\ifuniqueAffiliation % Standard variant of author block
\else
\usepackage{authblk}

\newbox{\orcid}\sbox{\orcid}{\includegraphics[scale=0.06]{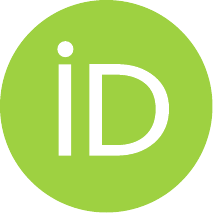}} 
\author[1]{%
	\href{https://orcid.org/0000-0002-7179-3664}{\usebox{\orcid}\hspace{1mm}Thorsten Hoeser}%
	\thanks{Corresponding author \texttt{thorsten.hoeser@dlr.de}}%
}

\author[1]{%
	\href{https://orcid.org/0000-0001-6181-0187}{\usebox{\orcid}\hspace{1mm}Felix Bachofer}%
}

\author[1,2]{%
	Claudia Kuenzer%
}

\affil[1]{Earth Observation Center (EOC), German Aerospace Center (DLR), Oberpfaffenhofen, 82234 Wessling, Germany}

\affil[2]{Institute for Geography and Geology, University of Wuerzburg, 97074 Wuerzburg, Germany}
\fi

\renewcommand{\headeright}{ }
\renewcommand{\undertitle}{Preprint}
\hypersetup{
pdftitle={Global Offshore Wind Infrastructure: Deployment and Operational Dynamics from Dense Sentinel-1 Time Series},
pdfauthor={Thorsten~Hoeser, Felix~Bachofer, Claudia~Kuenzer},
}

\begin{document}
\maketitle
\begin{abstract}
The offshore wind energy sector is expanding rapidly, increasing the need for independent, high-temporal-resolution monitoring of infrastructure deployment and operation at global scale. While Earth Observation based offshore wind infrastructure mapping has matured for spatial localization, existing open data sets lack temporally dense and semantically fine-grained information on construction and operational dynamics. We introduce a global Sentinel-1 synthetic aperture radar (SAR) time series data corpus that resolves deployment and operational phases of offshore wind infrastructure from 2016Q1 to 2025Q1. Building on an updated object detection workflow, we compile 15,606 time series at detected infrastructure locations, with overall 14,840,637 events as analysis-ready 1D SAR backscatter profiles, one profile per Sentinel-1 acquisition and location. To enable immediate use and benchmarking, we release (i) the analysis ready 1D SAR profiles, (ii) event-level baseline semantic labels generated by a rule-based classifier, and (iii) an expert-annotated benchmark data set of 553 time series with 328,657 event labels. The baseline classifier achieves a macro F1 score of 0.84 in event-wise evaluation and an area under the collapsed edit similarity-quality threshold curve (AUC) of 0.785, indicating temporal coherence. We demonstrate that the resulting corpus supports global-scale analyses of deployment dynamics, the identification of differences in regional deployment patterns, vessel interactions, and operational events, and provides a reference for developing and comparing time series classification methods for offshore wind infrastructure monitoring.
\end{abstract}

% keywords can be removed
\keywords{Earth Observation \and Offshore Wind Energy \and Offshore Infrastructure \and Sentinel-1 \and Time Series}

%\linenumbers                % Enable line numbering
\section{Introduction}
\label{sec:introduction}
\begin{figure}
	\centering
	\includegraphics[width=\linewidth]{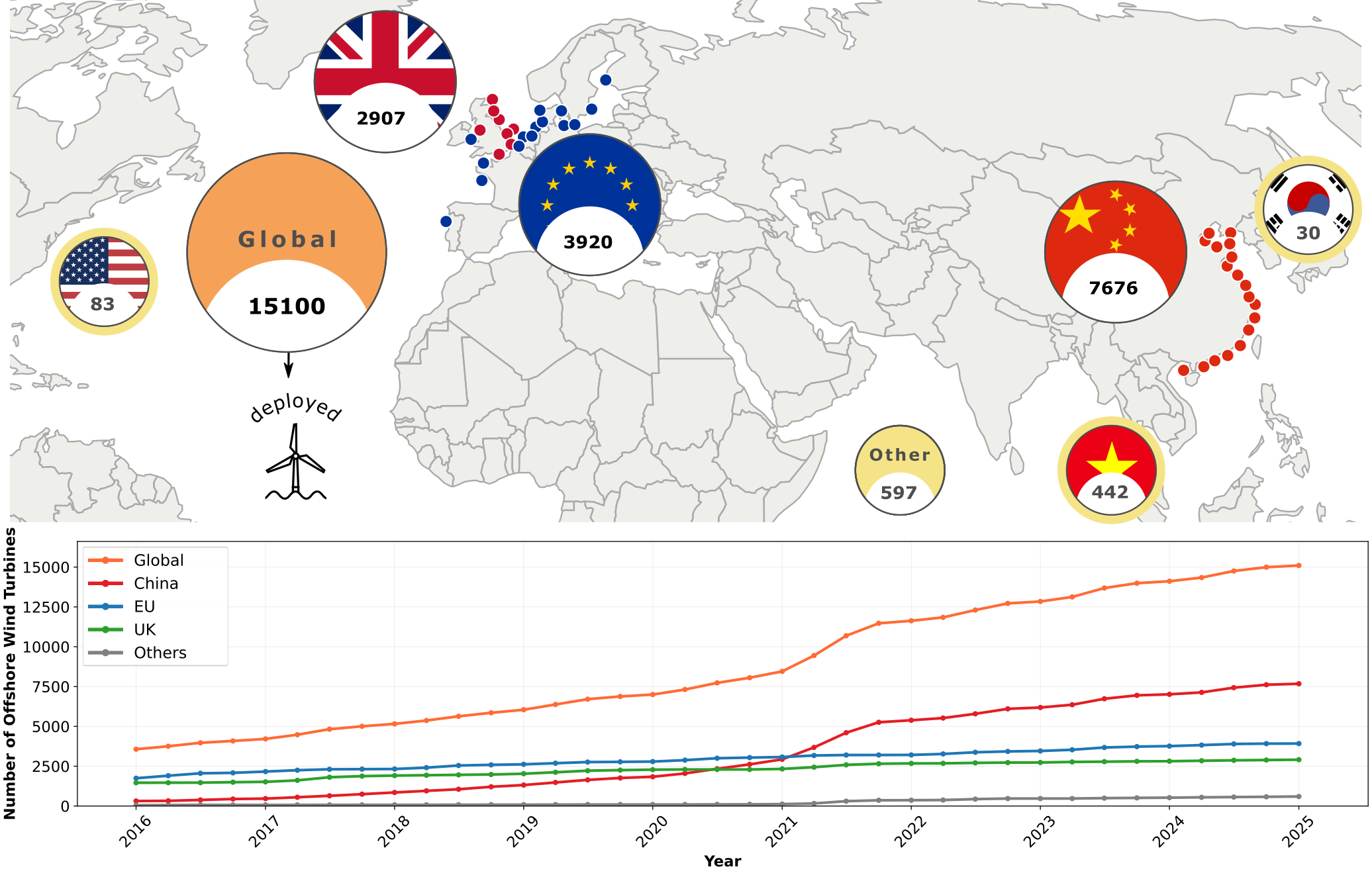}
	\caption{Major offshore wind energy markets with the number of readily deployed turbines in the first quarter of 2025 and their corresponding historic evolution aggregated to quarterly sums.}
	\label{fig:n_turbine_evolution}
\end{figure}

Offshore wind energy has experienced its most substantial growth phase within the past five years, see Figure \ref{fig:n_turbine_evolution}. The year 2021 stands out as a record year, during which 3,418 offshore wind turbines (OWTs) were deployed, the highest annual number to date. For comparison, the median annual deployment between 2017 and 2024 was 1,100 units, with the second highest year being 2023 (1,269 units). Regional differences show that the surge in 2021 was driven by China, which accounted for 77.3\% (2,642 units) of global deployments. The United Kingdom contributed 10.5\% (359 units), and the European Union 4.8\% (163 units). As of March 2025, the offshore wind market remains concentrated in these three regions: China accounts for 50.8\% (7,676 units) of all globally installed OWTs (15,100 total), followed by the European Union with 26\% (3,919 units) and the United Kingdom with 19.3\% (2,907 units).

This trend is expected to continue. Germany and the United Kingdom have set targets of 30 GW \citep{WindSeeG2016} and 40 GW \citep{UKGov2021}, respectively, by 2030. Looking further ahead to 2050, the European Union aims to reach 300 GW of offshore wind energy installed capacity \citep{EC2020}. China plans to expand its combined installed solar and wind energy capacity to six times its 2020 level, reaching approximately 3,600 GW \citep{UNClimateSummit2025}, with offshore wind representing a rapidly growing component in this strategy. Considering these targets, the recent surge in offshore wind infrastructure deployments marks only the beginning of a much larger expansion over the coming decades with a regional focus in the North Sea Basin and along the coast of China.

Beyond established markets, additional regions are expected to contribute to future expansion. Emerging technologies, particularly floating offshore wind turbines successfully demonstrated in pilot projects over the past decade, enable deployment in deeper waters and open unused marine areas for development \citep{10.1063/1.3435339, BENTO201966, hoeser2022heightcapa}. Today the offshore wind energy sector is characterized by both continued densification of established hotspot regions and the expansion into new offshore areas. These parallel developments require systematic monitoring of infrastructure deployments and operations, supported by freely available, high-quality data for stakeholders across industry, policy, and science.

From a lifecycle perspective, the construction phase of offshore wind infrastructure is the most resource-intensive, logistically complex, and environmentally impactful stage. OWT deployment is itself a multi-stage process. Site investigation vessels visit a location prior to construction, followed by foundation construction and transition piece installation, then pole, nacelle, and rotor blade installation, and finally continued operation with inspection and maintenance, which are indicated by vessel interaction or, in the case of floating turbines, towing events and the absence of the entire installation. Each stage has individual duration ranging from days to months, related by aspects such as project planning, turbine type, and site. The transitions between stages are typically where the impact at the site occurs and these transition events typically happen within days to weeks. Thus, high-density time series are necessary to provide narrow temporal bounds within which events such as site investigations by research vessels, monopile foundation construction, pole and nacelle installation or mooring events during operation due to maintenance can be isolated.

Only mapping binary turbine - no-turbine changes or providing changes between stages based on temporally aggregated signals (e.g. monthly or quarterly) underestimate the real complexity of an OWT deployment process. Improved understanding and monitoring of individual stages in the OWT lifecycle are essential for optimizing deployment and assessing impacts on marine activities and environments. Capturing and resolving these nuanced events with EO data requires dense temporal time series.

In this publication we present an approach and data set which captures single events by investigating high-density global time series of Sentinel-1 radar Earth Observation data. The primary objective is to create a consistent data set that allows for the analysis of time series and events at infrastructure locations. To this end, we present global Sentinel-1 SAR time series with median revisit intervals of up to one day within the European Union and 12 days outside the European Union, resolving deployment and operational phases of offshore wind infrastructure from 2016 to 2025. To our knowledge, there is currently no other study, which resolves multiple stages of offshore wind infrastructure deployment at this temporal resolution and doing this on a global scale with Earth Observation data.

The contributions of this paper are as follows:

\begin{itemize}
\item A global data set of 15,606 offshore wind infrastructure locations with global spatial coverage.
\item 14,840,637 SAR-based event signals compiled into 15,606 analysis ready time series corresponding to individual infrastructure locations spanning 2016Q1 to 2025Q1.
\item A rule-based baseline classification that assigns semantic labels to each event within all time series, distinguishing deployment and operational phases.
\item A benchmark data set consisting of 553 time series with 328,657 expert annotations at the event level, enabling evaluation and methodological comparison.
\end{itemize}

Collectively, these contributions establish an open, Earth Observation data corpus that combines global spatial coverage, high temporal resolution, analysis ready time series, baseline predictions, and an annotated benchmark data set \citep{hoeser_2026_18735421}. The proposed data set enables in-depth analysis of recent global offshore wind infrastructure dynamics and support the development, evaluation, and comparison of time series analysis and classification methods for offshore wind infrastructure applications.

\section{Related Research}
\label{sec:relatedreserach}

{\scriptsize
\setlength{\LTcapwidth}{\textwidth}
\begin{longtable}{@{}>{\RaggedRight}p{0.090\textwidth}>{\RaggedRight}p{0.090\textwidth}>{\RaggedRight}p{0.115\textwidth}>{\RaggedRight}p{0.110\textwidth}>{\RaggedRight}p{0.025\textwidth}>{\RaggedRight}p{0.025\textwidth}>{\RaggedRight}p{0.075\textwidth}>{\RaggedRight}p{0.090\textwidth}>{\RaggedRight}p{0.025\textwidth}>{\RaggedRight}p{0.025\textwidth}>{\RaggedRight}p{0.025\textwidth}>{\RaggedRight}p{0.025\textwidth}>{\RaggedRight}p{0.025\textwidth}>{\RaggedRight}p{0.025\textwidth}@{}}
  \caption{Overview of offshore wind turbine and infrastructure detection studies. The right-hand block reports which classes appear as explicit output labels of each method. The column-header icons stand for: \iconOpenWater{}~open water, \iconVesselSite{}~vessel at site, \iconTurbine{}~operational wind turbine, \iconPlatform{}~non-turbine offshore platform (oil/gas rig, substation, met mast), \iconFoundation{}~foundation / under-construction stage, \iconMooring{}~mooring construction or maintenance vessel. RGB/MS refers to optical or multispectral data. The column \textit{Output Granularity} contains short descriptions: \textit{Location only} means, that only spatial locations are provided but no temporal dynamics; \textit{Binary at e.g. date} means that class -- no-class binary labels have been derived, but no transition between construction phases; \textit{Multi-stage at e.g. date} means that the deployment process is not binary but resolved into multiple stages.} \label{tab:review} \\
  \toprule
  \textbf{Study} & \textbf{Method} & \textbf{Site} & \textbf{Sensor} & \rotatebox{90}{\textbf{Radar}} & \rotatebox{90}{\textbf{RGB/MS}} & \textbf{Temp. Range} & \textbf{Output Granularity} & \iconOpenWater & \iconVesselSite & \iconTurbine & \iconPlatform & \iconFoundation & \iconMooring \\
  \midrule
  \endfirsthead
 
  \multicolumn{14}{@{}l}{\footnotesize\textit{(continued from previous page)}} \\
  \toprule
  \textbf{Study} & \textbf{Method} & \textbf{Site} & \textbf{Sensor} & \rotatebox{90}{\textbf{Radar}} & \rotatebox{90}{\textbf{RGB/MS}} & \textbf{Temp. Range} & \textbf{Output Granularity} & \iconOpenWater & \iconVesselSite & \iconTurbine & \iconPlatform & \iconFoundation & \iconMooring \\
  \midrule
  \endhead
 
  \midrule
  \multicolumn{14}{r@{}}{\footnotesize\textit{(continued on next page)}} \\
  \endfoot
 
  \bottomrule
  \endlastfoot
  
  \citet{LIU201673} & Rule based, Statistical model & Regional: Gulf of Thailand, Gulf of Mexico and Persian Gulf & Landsat-8 & \xmark & \cmark & 2013-04 -- 2015-03 & Location only & \xmark & \xmark & \xmark & \cmark & \xmark & \xmark \\
  \midrule
  \citet{WONG2019111412} & CFAR, Rule based & Regional: Gulf of Mexico, UK and China & Sentinel-1 & \cmark & \xmark & 2017 & Location only & \xmark & \xmark & \cmark & \cmark & \xmark & \xmark \\
  \midrule
  \citet{s19020231} & CFAR, Rule based & Sub-national: \textit{South China Sea}, China & GaoFen-3 & \cmark & \xmark & 2017-09 & Location only & \xmark & \xmark & \xmark & \cmark & \xmark & \xmark \\
  \midrule
  \citet{e21060556} & CFAR, Statistical model & Sub-national: \textit{South China Sea}, China & RadarSat-2 & \cmark & \xmark & 2014-03 -- 2014-10 & Location only & \xmark & \xmark & \xmark & \cmark & \xmark & \xmark \\
  \midrule
  \citet{he2025onoffshore} & Rule based, Statistical model & National: China & Sentinel-2 & \xmark & \cmark & 2023 & Location only & \xmark & \xmark & \cmark & \xmark & \xmark & \xmark \\
  \midrule
  \citet{fei2025yolov5cdb} & Deep Learning & Global (sparse) & various satellite and aerial imagery & \xmark & \cmark & \textasciitilde{}2020-2024 (mosaic dates) & Location only & \xmark & \xmark & \cmark & \xmark & \xmark & \xmark \\
  \midrule
  \citet{song2025seanson} & Deep Learning & National: China & Sentinel-2 & \xmark & \cmark & 2023-01 -- 2025-03 & Location only & \xmark & \xmark & \cmark & \xmark & \xmark & \xmark \\
  \midrule
  \citet{Ding2024owtcn} & Deep Learning & National: China & Sentinel-1 & \cmark & \xmark & 2015-12 -- 2022-12 & Binary at annual frequency & \xmark & \xmark & \cmark & \xmark & \xmark & \xmark \\
  \midrule  
  \citet{XU2020110167} & Rule based, Statistical model & Regional: North Sea and surrounding waters & Landsat-5, Landsat-7, Landsat-8, Sentinel-2 & \xmark & \cmark & 2008-03 -- 2018-07 & Binary at date & \xmark & \xmark & \cmark & \xmark & \xmark & \xmark \\
  \midrule
  \citet{wang2024owtchina} & Rule based, Statistical model & National: China & Sentinel-1, Sentinel-2 & \cmark & \cmark & 2018 -- 2022 & Binary at annual frequency & \xmark & \xmark & \cmark & \xmark & \xmark & \xmark \\
  \midrule
  \citet{liu2024shandong} & Deep Learning & Sub-national: Shandong Sea, China & Sentinel-2 & \xmark & \cmark & 2021-03 -- 2023-03 & Binary at date & \xmark & \xmark & \cmark & \xmark & \xmark & \xmark \\
  \midrule
  \citet{zhang2021gowt} & Rule based, Statistical model & Global & Sentinel-1 & \cmark & \xmark & 2015-01 -- 2019-12 & Binary at annual aggregates & \xmark & \xmark & \cmark & \xmark & \xmark & \xmark \\
  \midrule
  \citet{WANG2024explosivegrowth} & Rule based, Statistical model & Global & Sentinel-1, Landsat-8 & \cmark & \cmark & 2000 -- 2022 & Binary at annual aggregates & \xmark & \xmark & \cmark & \xmark & \xmark & \xmark \\
  \midrule
  \citet{Zhang2024gowtgeedeep} & Deep Learning & Global & Sentinel-1, Sentinel-2 & \cmark & \cmark & 2014-10 -- 2023-12 & Binary at monthly aggregates & \xmark & \xmark & \cmark & \xmark & \xmark & \xmark \\
  \midrule
  \citet{LIU2026108706} & Deep Learning & National: China & Sentinel-1 & \cmark & \xmark & 2015-12 -- 2023-12 & Binary at annual frequency & \xmark & \xmark & \cmark & \xmark & \xmark & \xmark \\
  \midrule
  \citet{Paolo2024} & CFAR, Deep Learning & Global & Sentinel-1, Sentinel-2 & \cmark & \cmark & 2016-10 -- 2022-02 / (ongoing) & Binary at monthly frequency & \xmark & \cmark & \cmark & \cmark & \xmark & \xmark \\
  \midrule
  \citet{hoeser2022deepowt} & Deep Learning & Global & Sentinel-1 & \cmark & \xmark & 2016-07 -- 2021-06 & Multi-stage at quarterly aggregates & \cmark & \xmark & \cmark & \cmark & \cmark & \xmark \\
  \midrule
  \textit{ours} & Rule based, Deep Learning & Global & Sentinel-1 & \cmark & \xmark & 2016-01 -- 2025-03 & Multi-stage at date & \cmark & \cmark & \cmark & \cmark & \cmark & \cmark \\
\end{longtable}
}

The availability of global location data for offshore wind infrastructure has improved in recent years, particularly through Earth Observation-derived products. Table \ref{tab:review} provides an overview of studies and their contributions to offshore (wind) infrastructure mapping. The field has evolved progressively, starting from general offshore platform detection \citep{LIU201673, s19020231, e21060556}, diversifying platform classes to differentiate OWTs from other platforms \citep{WONG2019111412}, and increasing the spatial extent from regional studies to global applications \citep{zhang2021gowt, hoeser2022deepowt, WANG2024explosivegrowth, Zhang2024gowtgeedeep, Paolo2024}. The methods employed and sensors investigated are not yet consolidated. However, global studies tend to rely on Sentinel-1 radar data \citep{zhang2021gowt, hoeser2022deepowt}, which are also complemented by Sentinel-2 multispectral imagery \citep{WANG2024explosivegrowth, Zhang2024gowtgeedeep, Paolo2024}.

A key insight from this overview is the relationship between the temporal component and the output granularity. Currently, only binary no-turbine/turbine information is provided at the single-date level, allowing the monitoring of when persistent objects appear at an OWT site (see studies labeled \textit{Binary at date} in the \textit{Output Granularity} column of Table \ref{tab:review}). However, this binary information does not resolve the multi-stage deployment process occurring at an OWT location, which is responsible for the greatest engineering and logistical challenges, as well as the most significant impacts on marine environments, as stated in Section \ref{sec:introduction}. The only recent study attempting to resolve the deployment process uses quarterly aggregated SAR imagery \citep{hoeser2022deepowt}, which is not always able to temporally resolve major stage transitions and events occurring during deployment. In comparison, in this study we demonstrate how to provide multi-stage event labels at date, meaning, instead of aggregated SAR information, each individual acquisition is examined and events are reported on the date they were sensed. This provides the highest temporal resolution possible for the investigated sensor while delivering the semantically richest information to date compared to existing studies in the field.

The identified research gap, is a consequence of progressive maturity of OWT mapping happening over the last decade. Earth Observation based OWT mapping evolved consistently, with central studies each providing new aspects and refinements to the approach and demonstrated the capabilities of Earth Observation data in mapping marine infrastructure. \citet{LIU201673} used Landsat-8 imagery and identified offshore platforms in general by exploiting temporal and positional invariance of detected objects. \citet{XU2020110167} analyzed Sentinel-2 and Landsat data, applying statistical filtering and thresholding followed by spatial selection within reported wind farm areas in the North Sea Basin to derive OWT locations. Both studies contribute by demonstrating that OWTs and other marine infrastructures can be detected in multispectral imagery with a spatial resolution ranging from 10~m to 30~m.

In parallel, further studies demonstrated the suitability of SAR data for OWT detection. \citet{s19020231}, \citet{e21060556}, and \citet{WONG2019111412} applied constant false alarm rate (CFAR) to detect OWT locations in SAR imagery, leveraging the strong radar backscatter characteristics of turbine structures. These works laid an important foundation for SAR based OWT detection, by demonstrating this applicability.

Building on this foundation, the following studies contribute to OWT mapping by providing solutions which present results on a global scale. \citet{zhang2021gowt} applied morphological filtering to Sentinel-1 SAR imagery within a rule-based approach and produced a first freely available, Earth Observation based, global scale OWT data set called GOWF. A similar rule-based approach was presented by \citet{WANG2024explosivegrowth} also for global OWT location mapping, where the rule-set has been refined to provide robust decision boundaries for global applications. \citet{Paolo2024}, motivated by the work of \citet{WONG2019111412}, used a combination of CFAR on Sentinel-1 SAR data and convolutional neural network (CNN) on Sentinel-1 SAR and Sentinel-2 multispectral data to detect and classify marine infrastructure with a dedicated "wind" category. This study demonstrates, how the combination of multispectral and SAR imagery, processed by consecutive CFAR and deep learning analysis can scale to global marine infrastructure mapping. Another study by \citet{Zhang2024gowtgeedeep} focuses on precision maximization, especially important for near coast and heterogeneous coastal environments by investigating Sentinel-2 imagery with an ensemble of CNN based detection models and successfully reduced false positives in global OWT detection. \citet{hoeser2022deepowt} introduced DeepOWT, a data set of offshore wind infrastructure locations which beside its global scope is a first contribution to disentangle stages in the deployment process of OWTs, distinguishing OWT foundation, and fully deployed OWTs as separate classes, using a CNN based object detection approach on Sentinel-1 SAR data. 

From both the overview table as well as the more detailed review of central studies to the field, it becomes clear that OWT position mapping has been the major focus of the preceding studies. Now, for this study, the temporal dimension of offshore wind infrastructure deployment is of particular importance. Existing approaches that incorporate temporal information can broadly be grouped into two categories. The first category investigates binary changes in SAR amplitude or multispectral reflectance time series to detect persistent changes indicating turbine installation. The studies of \citet{XU2020110167}, \citet{zhang2021gowt}, \citet{Ding2024owtcn}, \citet{Paolo2024}, and \citet{LIU2026108706} fall into this category. These approaches provide binary signals (non-turbine/turbine) with annual to monthly temporal resolution. While effective for estimating the onset of deployment, they do not differentiate fine-grained construction stages such as foundation installation, tower and nacelle assembly, or periods of active construction.

The second category analyzes spatial SAR backscatter patterns to distinguish multiple deployment stages. \citet{hoeser2022deepowt} introduced this category by using three-month aggregated Sentinel-1 SAR data to differentiate between no turbine, turbine foundation, and deployed turbine stages. However, the quarterly aggregation reduces temporal precision and limits the ability to capture short-term events such as vessel interactions or precise phase transitions such as from open water to completed foundation, or from foundation to fully installed turbine. 

To address these limitations, we aim to establish a foundation for high-density time series analysis of offshore wind infrastructure, enabling temporally precise and semantically rich monitoring at a global scale. Such enriched data sets support market analyses and strategic outlooks \citep{JUNG2023popertiesowtglobal, JUNG2024futureglobalowt}, enable investigation of regional deployment and decommissioning dynamics \citep{GOURVENEC2022111794}, provide detailed inputs necessary for spatial planning and modelling \citep{SCHWARTZBELKIN2023106280}, and support ecosystem impact studies through precise temporal correlation of construction phases with animal behavior \citep{lai2024endangered}.

\section{Data and Materials}
\label{sec:data}

This study uses data from the Sentinel-1 satellite constellation, part of the European Space Agency’s (ESA) Copernicus Programme. Sentinel-1 provides C-band (5.6~cm wavelength) SAR imagery which is largely independent of cloud cover and solar illumination \citep{TORRES20129}. Combined with revisit intervals ranging from 1 to 12 days, depending on number of platforms, latitude and geographic region, Sentinel-1 is well suited for generating high-density time series at global scale, particularly for coastal and offshore regions.

\begin{figure}
	\centering
	\includegraphics[width=\linewidth]{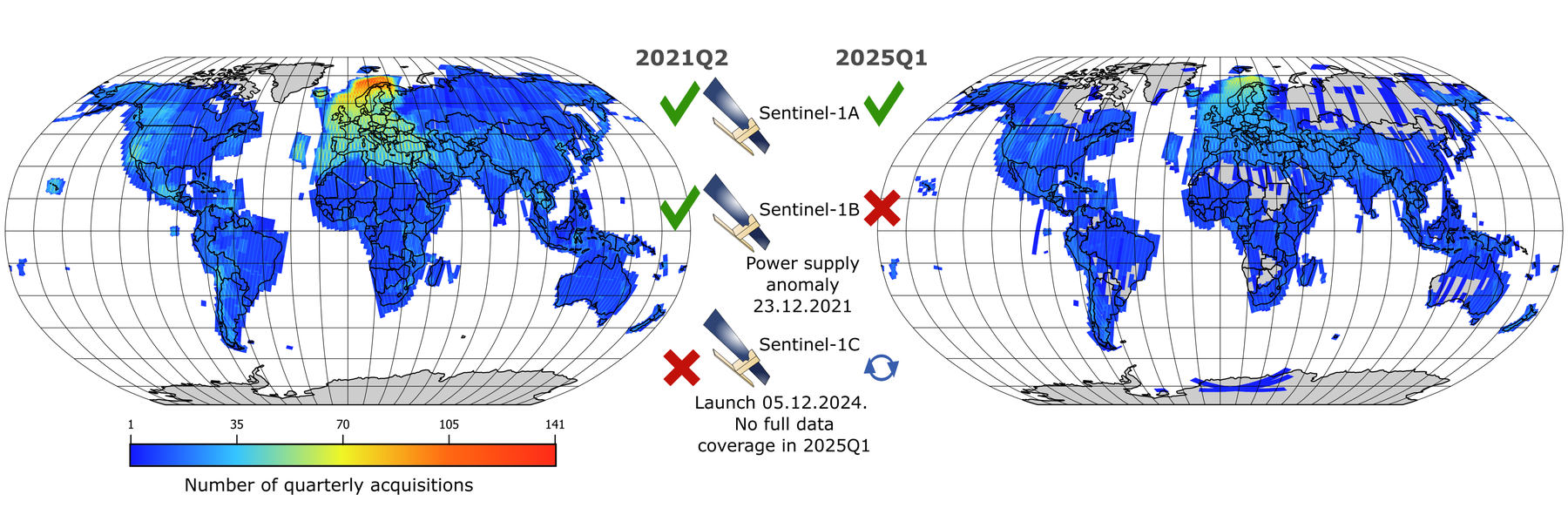}
	\caption{Global distribution of the number of quarterly acquisitions by the Sentinel-1 mission with different platforms in orbit and actively operating.}
	\label{fig:coverage}
\end{figure}

Spatial and temporal coverage highly depend on the number of operational platforms within the constellation, see Figure \ref{fig:coverage}. During the second quarter of 2021, both Sentinel-1A and Sentinel-1B were fully operational, with the highest data take frequency over the European Union. Following the power supply anomaly of Sentinel-1B on 2021-12-23 \citep{ESA_Sentinel1B_AnomalyReport}, and prior to the operational phase of Sentinel-1C (launched but not acquiring data in the first quarter of 2025), Sentinel-1A was the only operating platform during this time. Even when this resulted in reduced spatial coverage and lower revisit frequency, coastal and offshore regions continued to receive consistent acquisitions. The constellation’s fallback design demonstrates that even under an unplanned platform failure, Sentinel-1 remains capable of delivering continuous time series suitable for coastal and offshore monitoring applications.

\begin{figure}
	\centering
	\includegraphics[width=\linewidth]{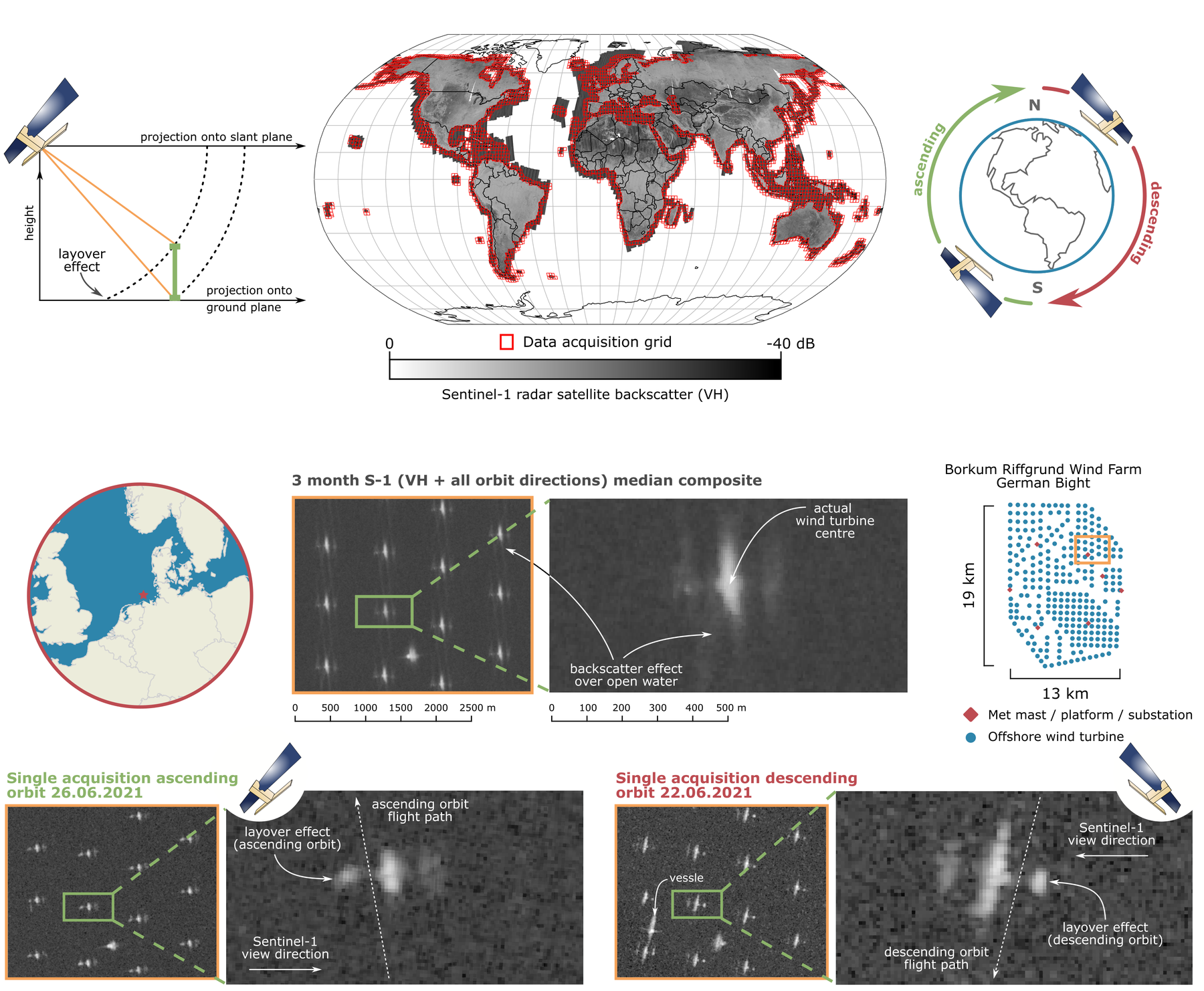}
	\caption{Global processing grid overlaid on a three-month Sentinel-1 median composite, together with schematics of the satellite's polar orbit and radar acquisition geometry and representative Sentinel-1 scenes demonstrating viewing-direction-dependent imaging effects.}
	\label{fig:sar_mechanics}
\end{figure}

For persistent offshore infrastructure detection, SAR backscatter characteristics are particularly suitable, see Figure \ref{fig:sar_mechanics}. Due to the near-polar orbit and right-looking acquisition geometry of Sentinel-1, images are acquired in ascending and descending passes, meaning targets are observed either from west-to-east or east-to-west viewing directions. For vertical metallic structures such as offshore wind turbines, this geometry produces characteristic radar signatures \citep{hoeser2022heightcapa}.

The layover effect creates such a characteristic signature, which occurs when the signal reflected from elevated structures returns to the sensor before the signal reflected from the ground or sea surface at the structure’s base. For OWTs, strong reflections from the nacelle generate high backscatter values that appear displaced toward the sensor, relative to the true turbine location. Visually, the turbine appears to lean toward the sensor. Depending on orbit direction, this layover signature appears to the left for ascending passes and right for descending passes relative to the OWT center. This horizontal signature is a key element of the data set design and subsequent analysis presented in this study. 

When combining acquisitions from different orbit directions into temporal composites like a median stack, the turbine center and layover-related backscatter peaks are sharpened, while speckle over open water is reduced. Transient objects such as vessels are largely suppressed in median composites, provided they do not remain stationary during the compositing period \citep{hoeser2022deepowt, Paolo2024}.

Polarization selection for the data used in this study was motivated by using the polarization which provide the highest contrast between targets (offshore infrastructures) and non-targets (open water). Open water shows very low backscatter in VH (vertical transmit, horizontal receive) polarization \citet{GARG2024114417}, while offshore wind turbines, due to their angular construction, still produce a clear VH signal through multiple scattering. Especially in water with higher turbidity, VH polarization suppresses signals from the water surface, e.g. variations coming from sediment plumes, while in VV (vertical transmit, vertical receive) polarization these signals are stronger and more variable \citet{Shao02112021}. Thus, VH polarization increases the contrast between water and the target objects relative to VH polarization, as investigated by \citet{rs9050440} and reported by \citet{s19020231}.

\begin{figure}
	\centering
	\includegraphics[width=0.8\linewidth]{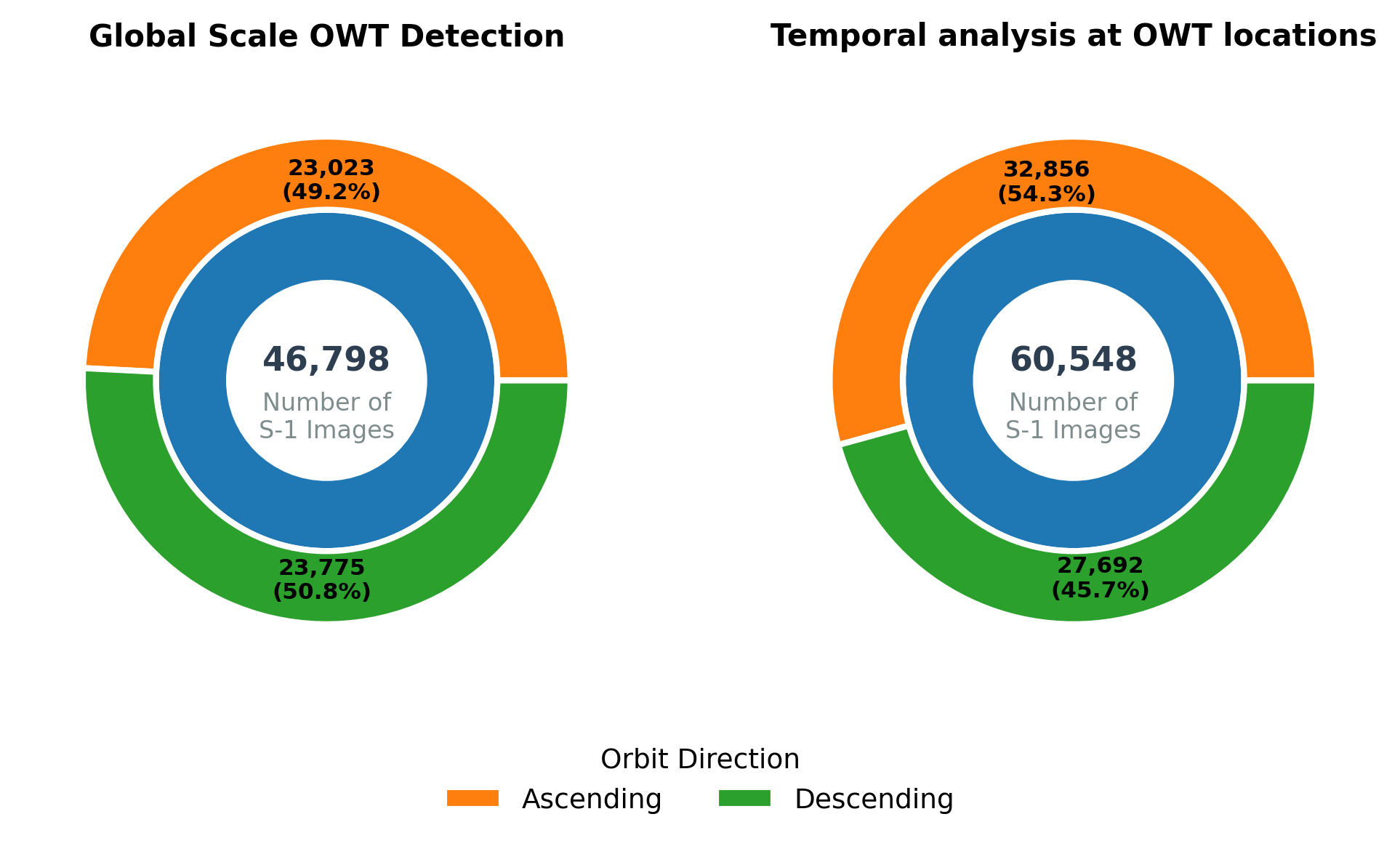}
	\caption{Number of Sentinel-1 images investigated in this study grouped by their respective process as explained in detail in Section \ref{section:method}.}
	\label{fig:used_data}
\end{figure}

Taking these properties of the underlying data into account, this study investigates Sentinel-1 Ground Range Detected (GRD) products acquired in Interferometric Wide (IW) swath mode with a ground sampling distance of 10~m, and VH polarization, preprocessed and made available via Google Earth Engine \citep{GORELICK201718, GEE_Sentinel1_Preprocessing}. We used both orbit directions, and all available platforms (Sentinel-1A and B) and did not limit the incident angle range to maximize the amount of images available for investigation. For OWT detection, we filtered images globally within a temporal period ranging from 2025-01-01 until 2025-03-31 covering the entire first quarter of the year and reduced this image collection to a median composite, removing non-stationary objects. By using the detected OWT locations from the median composite as spatial constraints, we then filtered the archive within a temporal period ranging from 2016-01-01 until 2025-03-31 to use each single acquisition per location to create the temporal signals. Figure \ref{fig:used_data} provides the summary statistics about the Sentinel-1 images used in this study.

\section{Method}
\label{section:method}
\subsection{Updated global offshore wind infrastructure detection}
\label{section:detection}

To derive offshore wind infrastructure locations at global scale, we updated the detection workflow proposed by \citet{hoeser2022deepowt}. The workflow consists of a global offshore wind farm proposal network, and a second-stage detector that identifies individual infrastructure units within the proposed regions.

\begin{figure}
	\centering
	\includegraphics[width=\linewidth]{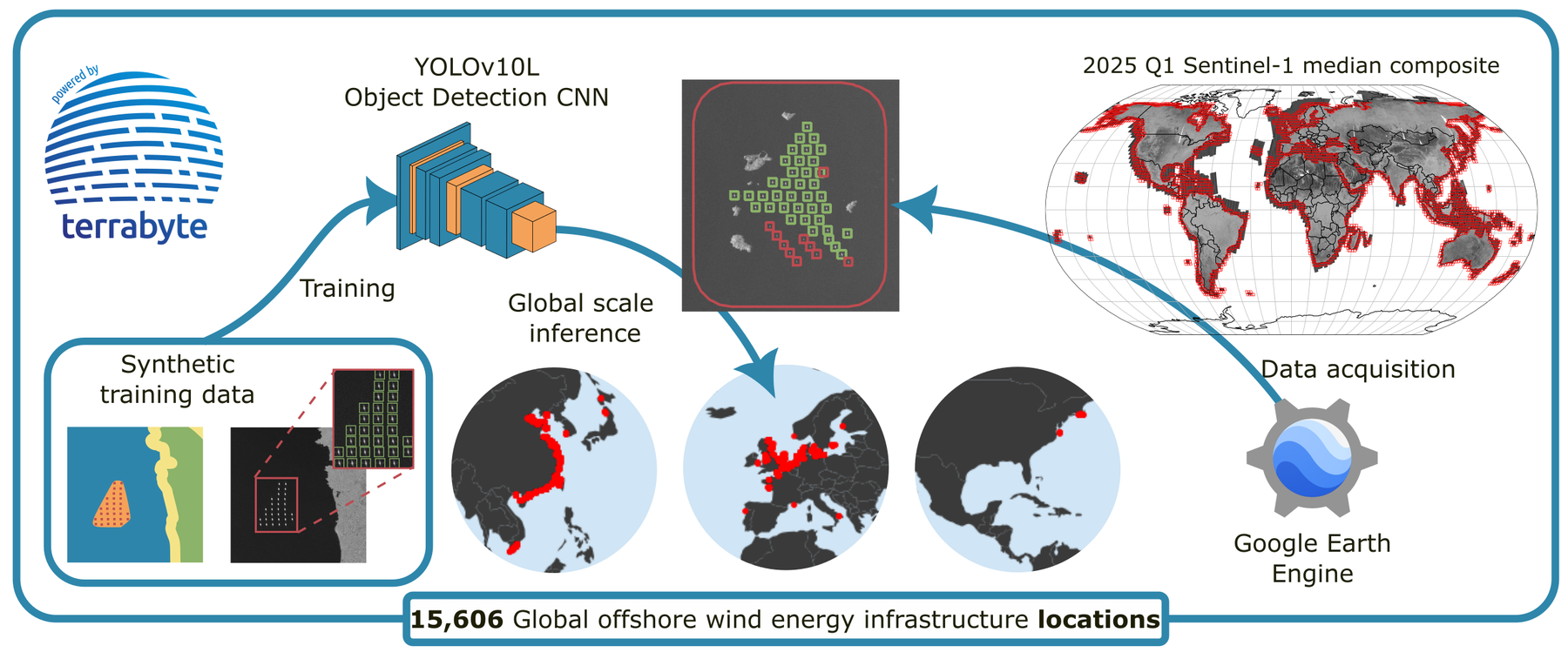}
	\caption{Deep learning based, global, offshore wind infrastructure detection workflow based on \citet{hoeser2022deepowt}.}
	\label{fig:detect_workflow}
\end{figure}

As illustrated in Figure \ref{fig:detect_workflow}, we trained YOLOv10-L object detection models with architecture and loss design provided by Ultralytics \citep{THU-MIGyolov10}. YOLOv10-L was chosen to improve inference throughput and to avoid non maximum suppression by using its one-to-one head architectural design during inference. It replaces the Faster R-CNN \citep{7485869} architecture used in the preceding study, improving cross validation site detection metrics for offshore wind infrastructure facilities as presented later in Figure \ref{fig:detection_performance}. Furthermore, the choice to move from a Faster R-CNN TensorFlow implementation \citep{hoeser2022deepowt} to models from the YOLO series maintained by Ultralytics was made to use recent and competitive detection models without the need to built a custom detector. In a study design whose methodological focus is not OWT location mapping but rather the classification of individual events in SAR backscatter time series, this choice frees resources from detection model architecture and training implementation, which, as recent studies demonstrate, requires considerable experimentation in the mature state OWT mapping is in \citep{LIU2026108706}. This in turn allows us to advance into the analysis of classifying single events at turbine locations in SAR time series.

For training of the YOLOv10-L models, we  followed \citet{HOESER2022synteo, hoeser2022deepowt} and generated 90,000 synthetic training samples with corresponding annotations for both offshore wind farms and individual offshore wind infrastructure units to optimize the models for wind farm and single unit detection. The training data sets contain images showing wind farms, offshore platforms which are not wind turbines, and images without any platforms showing only open sea or coastal environments. The ratio of these image types in the training data is 50/25/25 respectively and we split the entire set into a 80/20 train/val split during training. The training was conducted on four NVIDIA A100 GPUs hosted on DLR’s Terrabyte HPC \citep{terrabyte}. The models were trained using an input size of 1,024 × 1,024 pixels, a batch size of 32, and 20 epochs. We adapted the default training settings as reported in the models documentation \citep{jocher2026ultralyticscfg} as follows. A linearly decaying learning rate schedule was applied, decreasing from 0.01 to 0.0001 following a three-epoch warm-up phase, in combination with AdamW as optimizer. For data augmentation we used randomized vertical flipping, image scaling up to 20\%, and rotation between 0 and 180 degrees. 

Inference was performed on median composites of the Sentinel-1 GRD IW product. All images which are within the global Exclusive Economic Zones (EEZs) \citep{VLIZ2024} are queried using the filters described in Section \ref{sec:data} and reduced to a median composite on GEE for 2025Q1. The resulting tiles were exported to Terrabyte HPC and organized as a global mosaic using the GDAL tile index (GTI). A virtual chip grid of 2,048 $\times$ 2,048 pixels with 50\% overlap is used to enable highly parallelized data access via the GTI during a CPU based inference. We used a confidence threshold of 0.25 to keep predictions. Detected bounding boxes of the second object detection stage were reduced to centroid coordinates to get point representations of offshore wind infrastructures for validation.

Validation was performed using manually annotated offshore wind infrastructure locations derived from the Sentinel-1 2025Q1 median composite across three validation sites. We updated the validation data sets from \citet{hoeser2022deepowt} to include infrastructure constructed up to 2025Q1 in the North Sea Basin and the East China Sea. Additionally, a new validation site was introduced along the southeast coast of Vietnam to include challenging to detect offshore wind farm layouts characteristic for this region. Overall we provide 9,770 offshore wind energy infrastructure validation locations. Because classification of deployment state and object type is addressed in the subsequent time series analysis, detection was evaluated in a class-agnostic manner, treating turbine foundations, deployed turbines, and support platforms as a single class.

Performance was quantified using Precision, Recall, and F1 score at site level, along with macro- and micro-averaged scores across sites to account for class imbalance. A detection was considered a true positive (TP) if its centroid lay within a radius of 100~m of a ground truth point. Detections outside this radius were counted as false positives (FP), and ground truth points without a corresponding detection within 100~m were counted as false negatives (FN).

We handled edge cases, so that if multiple detections matched the same ground truth point, only the closest detection was counted as TP, additional detections were counted as FP. Furthermore, if a detection fell within 100~m of multiple ground truth points, it was assigned to the nearest ground truth location as TP and unmatched ground truth points were counted as FN.

Evaluation metrics were computed as:
\begin{equation}
\mathrm{Precision} = \frac{\mathrm{TP}}{\mathrm{TP} + \mathrm{FP}},
\end{equation}

\begin{equation}
\mathrm{Recall} = \frac{\mathrm{TP}}{\mathrm{TP} + \mathrm{FN}},
\end{equation}

\begin{equation}
\mathrm{F}_1 = 2 \times \frac{\mathrm{Precision} \times \mathrm{Recall}}{\mathrm{Precision} + \mathrm{Recall}}.
\end{equation}

With this modernized workflow building upon \citet{hoeser2022deepowt}, we provide an updated global data set of offshore wind infrastructure locations, which serves as the foundation for compiling and analyzing high-density time series of offshore wind infrastructure dynamics \citet{hoeser_2026_18735421}.

\subsection{High density Sentinel-1 time series compilation}
\label{section:compilation}

The primary contribution of this publication is the extension of global offshore wind infrastructure detections with high-density Sentinel-1 based time series, providing fine-grained details of temporally resolved, Earth Observation based information on deployment and operational phases of offshore wind infrastructure. To this end, we acquire and preprocess every available Sentinel-1 acquisition at each detected offshore wind infrastructure location individually.

\begin{figure}
	\centering
	\includegraphics[width=0.9\linewidth]{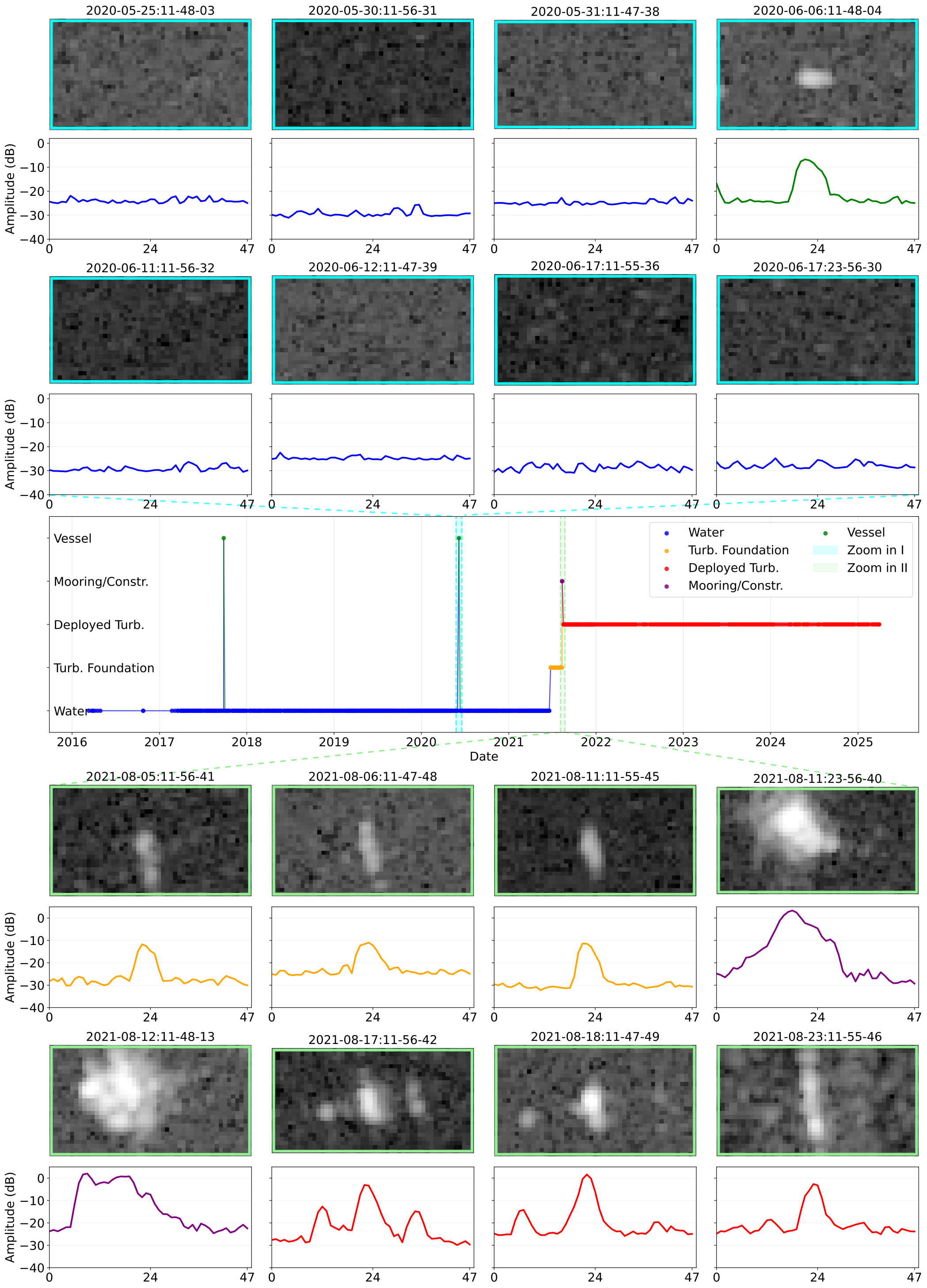}
	\caption{Example time series of a single offshore wind turbine. The central panel shows predicted event labels, the upper and lower panels display corresponding 2D Sentinel-1 backscatter images and derived 1D backscatter profiles.}
	\label{fig:temp_example}
\end{figure}

Figure \ref{fig:temp_example} illustrates a final example time series for a single offshore wind turbine covering the period from 2016 to 2025. The 1D backscatter profiles are generated by applying a column-wise maximum reduction along the horizontal axis of each 2D image. These profiles capture characteristic SAR signatures whose spatial patterns vary depending on the infrastructure state. The resulting time series consist of one 1D backscatter profile per Sentinel-1 acquisition per turbine location. The upper part of the figure shows a typical pre-construction phase with open water. On 2020-06-06, a survey vessel is present at the turbine location. This vessel interaction can be detected through short-term changes in the 1D profile, showing a moderate but distinct amplitude spike visible for only one or two acquisitions, with the 1D profiles showing water both before and after. The second inset shows a typical transition phase at a turbine foundation, which is visited by a construction vessel between 2021-08-11 and 2021-08-21. During this period, the vessel moors at the site and installs the pole and nacelle, the latter of which can be seen fully deployed on 2021-08-17.

\begin{figure}
	\centering
	\includegraphics[width=\linewidth]{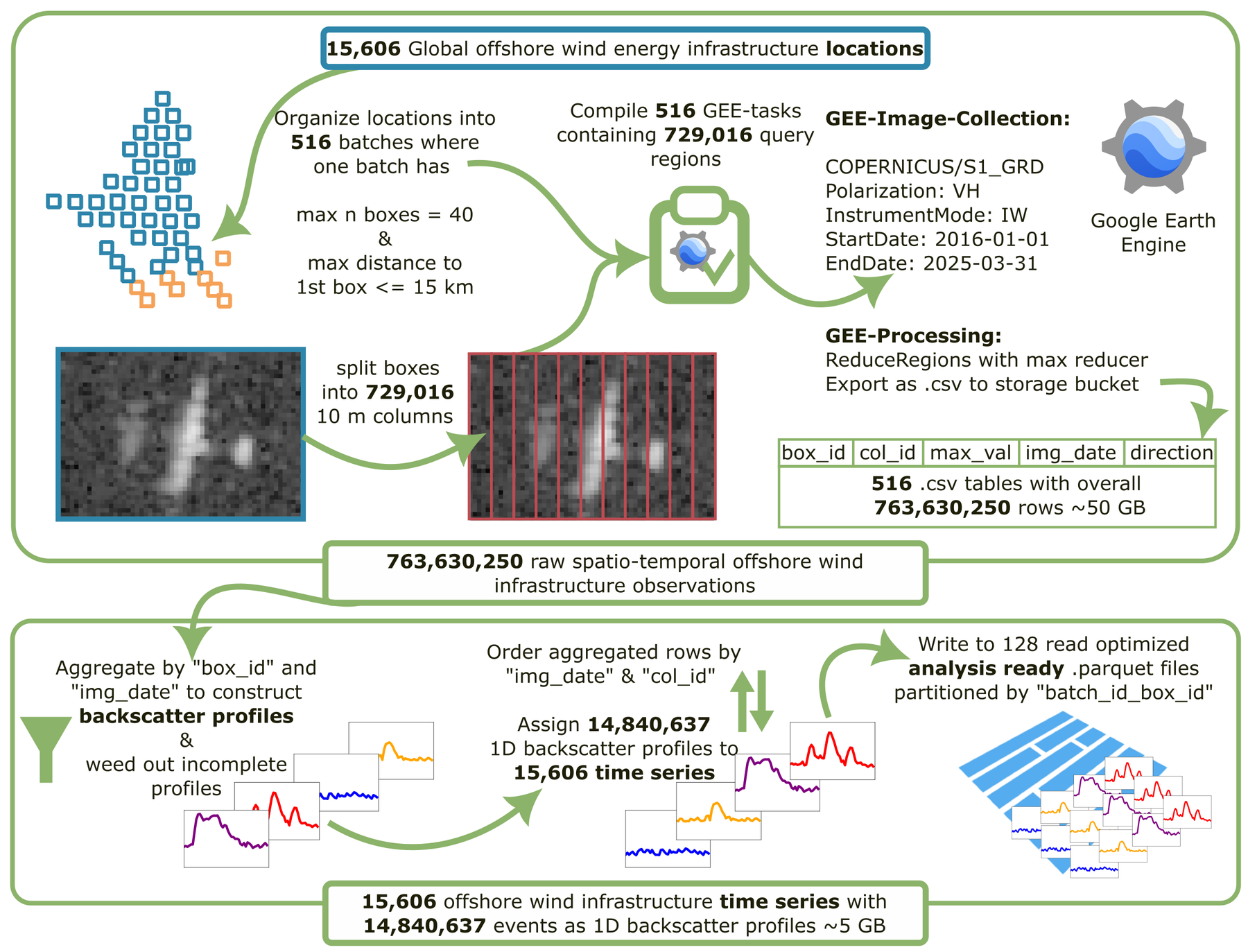}
	\caption{Time series compilation workflow, based on 15,606 offshore wind infrastructure spatial detections to efficiently acquire 14.8 million Sentinel-1, 1D backscatter profiles with unique time and location relations.}
	\label{fig:temp_workflow}
\end{figure}

Data acquisition was performed on GEE, while client-side postprocessing was used to compile the final analysis ready time series data set, see Figure \ref{fig:temp_workflow}. GEE strongly uses parallelizable map-reduce operations, where in our case, the applied operation consists of a column-wise maximum reduction to derive 1D profiles from 2D Sentinel-1 imagery crops.

Three main challenges had to be addressed:
\begin{enumerate}
\item Defining a spatial feature representation compatible with GEE’s region-based reduction.
\item Designing a scalable and efficient batch processing pipeline which matches GEE’s image collection filtering.
\item Implementing a data model compliant with GEE constraints while minimizing export size and data egress costs.
\end{enumerate}

A naïve approach would query an image, crop the region of interest, and apply a column-wise reduction to obtain a 1D profile. However, GEE’s region reduction assigns a single aggregated value per geometry, which would reduce each detection box into a single scalar rather than a profile.

To overcome this limitation, each of the 15,606 detection boxes was split client side into 10~m wide vertical columns along the horizontal axis, resulting in 729,016 narrow spatial processing units. Instead of launching one GEE task per column, boxes (not columns) were grouped into spatial batches under the following constraints:
\begin{itemize}
\item A batch includes all boxes within 15~km of a randomly selected seed box.
\item A batch contains no more than 40 boxes.
\end{itemize}

The 15~km spatial threshold increases the likelihood that all boxes in a batch share common image footprints, minimizing unnecessary NaN values due to partial coverage. The limit of 40 boxes per batch was determined empirically and maintains high parallelism while preventing overly large export objects that may cause GEE task failures. Applying this batching strategy resulted in 516 processing batches for the 15,606 detected locations.

Each GEE batch exported a CSV file. The tabular format was selected to allow inclusion of metadata alongside reduced backscatter values. The data model contains:
\begin{itemize}
\item \texttt{box\_id}: identifier of the parent detection box
\item \texttt{column\_id}: identifier of the 10~m column within the box
\item \texttt{max\_value}: maximum backscatter value for the column
\item \texttt{img\_date}: Sentinel-1 acquisition date
\item \texttt{orbit\_direction}: ascending or descending pass
\end{itemize}

Including orbit direction is essential for later interpretation of directional SAR signatures.

The 516 exported CSV files collectively contain 763,630,250 rows. GEE task preparation and submission was managed client side, maintaining up to 25 concurrent or queued GEE tasks with automated retry logic after failure. Data were exported to a Google Cloud Storage bucket, resulting in about 50 GB of data and total egress cost of about 6~USD. This cloud-based workflow demonstrates both computational and resource efficiency for large-scale SAR time series acquisition.

After download, batch level CSV files were processed individually. Rows were aggregated by \texttt{box\_id} and \texttt{img\_date} to group all backscatter profile values belonging to a single location and acquisition. Entries containing NaN values were removed. To reconstruct coherent 1D backscatter profiles, column values were sorted by \texttt{column\_id}. This resulted in 14,840,637 valid 1D backscatter profiles associated with offshore wind infrastructure locations and unique acquisition dates.

For efficient storage and data access, the data set was saved into 128 partitioned Parquet files. Partitioning was implemented by computing a hash of \texttt{batch\_id} + \texttt{box\_id} and mapping the resulting integer to the range 0-127 using a modulo operation. This scheme ensures that all profiles belonging to the same location are stored within the same partition while maintaining near-uniform partition sizes. As a result, loading individual location-specific time series is highly efficient.

\subsection{Rule-based time series classification}
\label{section:classifier}

The goal is to provide a rule-based classifier as baseline approach, which is able to assign the labels water, turbine foundation, deployed turbine, vessel, mooring / active construction, or as fallback unclear to each 1D backscatter profile of the prepared time series. Today, machine and deep learning approaches dominate most data-heavy research. However, expert-knowledge-based methods such as rule-based decision trees are still useful, since they are interpretable and computationally efficient, and they can be used as a reference to measure machine learning approaches. This makes it possible to quantify the performance gain of machine learning approaches and to judge whether their higher computational cost or lack of traceability is justified. Furthermore, in this case the choice of a rule-based approach is also empirically grounded, since the 1D swath profiles show clearly visible, class-specific patterns whose isolation is technically feasible to a specific extent, making an explainable rule-based classifier a methodologically sound first step before advancing to parameter-heavy solutions such as neural networks.

To prepare the 1D swath profiles to enter the rule-based classifier, for each backscatter profile peak detection is performed \citep{2020SciPyNMeth} after a Gaussian smoothing of the original backscatter profile to focus on major peaks. During peak detection, candidate peaks must satisfy a minimum inter-peak distance of 5 profile bins and a minimum prominence of 2~dB. The detected peaks are mapped back to their positions to the unsmoothed profile indices. To distinguish peaks resulting from the actual target position and layover effects, the center peak is defined as the peak closest to the profile center. Layover peaks are defined based on orbit direction, for descending passes, the right adjacent peak serves as the valid side peak, and for ascending passes, the left adjacent peak is used. Furthermore, for each profile we calculate the mean, standard deviation and range. This set of features is used during profile classification.

The rule-based classifier is designed in two stages. In the first stage, each profile is classified independently using thresholds on profile statistics (mean, standard deviation, range) and peak characteristics (presence, prominence, width, and amplitude). Figure \ref{fig:temp_example} should be revisited here to get an intuition of the characteristics of the 1D backscatter profiles and different profile types. A water label is assigned if the profile has low overall backscatter or lacks distinct peaks, including cases of low mean and variability, weak or absent center peaks, or uniformly low peak prominence. A mooring / active construction label would indicate the presence of a larger construction vessel or maintenance vessel, for which we assume strong and broad backscatter profiles. This label is assigned when broad peaks or high dynamic range indicate strong but spatially diffuse responses, or when low variance coincides with high mean backscatter. A deployed turbine label is assigned when a pronounced center peak and a valid layover peak are present with sufficient amplitude and prominence. An unclear label is assigned to profiles with elevated side peaks that do not satisfy any of the preceding criteria, while a turbine foundation label is assigned when a sufficiently prominent center peak is present and no earlier rule applies. All remaining profiles default to unclear after this first stage.

In the second stage, labels are refined using sequential context. First, a smoothing operation is performed by applying a moving window along the chronologically ordered class labels of the first stage with an adaptive kernel that ignores transient labels (unclear, vessel, mooring) when determining effective neighbors. Targeted passes correct isolated non-water profiles flanked by water and enforce local consistency by relabeling profiles that disagree with two agreeing neighbors, except for vessel labels, which are preserved. Next, profiles still labeled unclear are reevaluated using relaxed peak-based criteria to resolve ambiguities between mooring / active construction and turbine foundation classes. The smoothing step is then repeated to propagate updated labels.

Subsequent refinements correct isolated water labels between turbine-related classes using a sliding window. Larger temporal contexts are then considered through segment-based refinement, where adjacent deployed turbine and turbine foundation segments are compared chronologically. Treating mooring and vessel labels as transparent, the shorter of two conflicting adjacent segments is relabeled to match the longer one. This process is then iterated until convergence.

Finally, a timeseries refinement resolves platform related ambiguities by examining the global label distribution. Depending on the dominance and ordering of vessel, mooring, turbine foundation, and deployed turbine labels, subsets of these labels are reassigned to the platform class, including cases where turbine foundation labels persist long after the last deployed turbine occurrence. The final labels are written to the time series parquet files holding the corresponding 14,840,637 swath profiles.

To evaluate the performance of the rule-based classification, we examined and annotated 328,657 swath profiles of 553 time series randomly drawn from all 15,606 available time series. This benchmark data set is part of the data corpus provided with this publication. We used two validation strategies, single event classification evaluation and time series evaluation.

For single event evaluation, the ground-truth event labels in the benchmark data set are compared against the labels predicted by the rule-based classifier using a confusion matrix, from which class-wise Precision, Recall, and F1-scores are derived, as well as overall and grouped macro averaged scores.

To complement these point wise scores with a sequence sensitive metric, we compute the edit similarity score. Edit similarity is defined as the normalized Levenshtein distance \citep{levenshtein1966binary} between the ground-truth and predicted label sequences, which we chose due to different sequence lengths of the time series:

\begin{equation}
\mathrm{EditSim}(S, \hat{S}) = 1 - \frac{d_{\mathrm{Lev}}(S, \hat{S})}{\max(|S|, |\hat{S}|)},
\end{equation}

where $d_{\mathrm{Lev}}$ denotes the Levenshtein distance and $|S|$ and $|\hat{S}|$ are the sequence lengths. The Levenshtein distance measures how many edit operations like insertions, deletions, or substitutions of labels are required to align the predicted sequence with the ground-truth sequence, where fewer required edits relate to a higher similarity. Edit similarity is computed in two variants. The first variant compares the full label sequences performing a point-wise evaluation, thereby penalizing incorrect labels at any time step. Since this variant is conceptually close to a micro-averaged F1 score obtained from single-event evaluation, we additionally compute edit similarity on collapsed label sequences. During collapsing, segments of identical labels are reduced to a single label. Collapsed edit similarity abstracts away absolute event durations and focuses on the temporal ordering and presence of event phases. This variant penalizes unnecessary transitions, incorrect event ordering, and missing or extra phases in the predicted sequence, making it a suitable complementary metric to single-event evaluation. For benchmarking purposes, edit similarity is further evaluated at uniformly spaced quality thresholds $q_s \in [0, 1]$ with a step size of 0.05. For each threshold, we compute the fraction of time series with an edit similarity score greater than or equal to $q_s$. The overall performance is summarized as the area under the edit similarity-quality threshold curve (AUC).

\section{Results}
\label{section:results}

\subsection{Dense SAR backscatter profile time series}
\label{section:timesereis}

As a primary outcome of this study, we provide a globally consistent, time series data set of offshore wind infrastructure derived from Sentinel-1 SAR imagery available at \citet{hoeser_2026_18735421}. While an updated set of offshore wind infrastructure locations for 2025Q1 is included as a structural foundation, the central contribution lies in the analysis-ready time series data and the benchmark annotations.

The 14,840,637 analysis-ready 1D Sentinel-1 backscatter profiles spanning the period from 2016Q1 to 2025Q1 are a starting point for detailed investigation and research in Earth Observation based offshore wind infrastructure monitoring by complementing spatial detection approaches with temporal infrastructure dynamics. Each backscatter profile corresponds to a single acquisition at a detected offshore wind infrastructure location and preserves structurally informative SAR signatures.

\begin{figure}
	\centering
	\includegraphics[width=\linewidth]{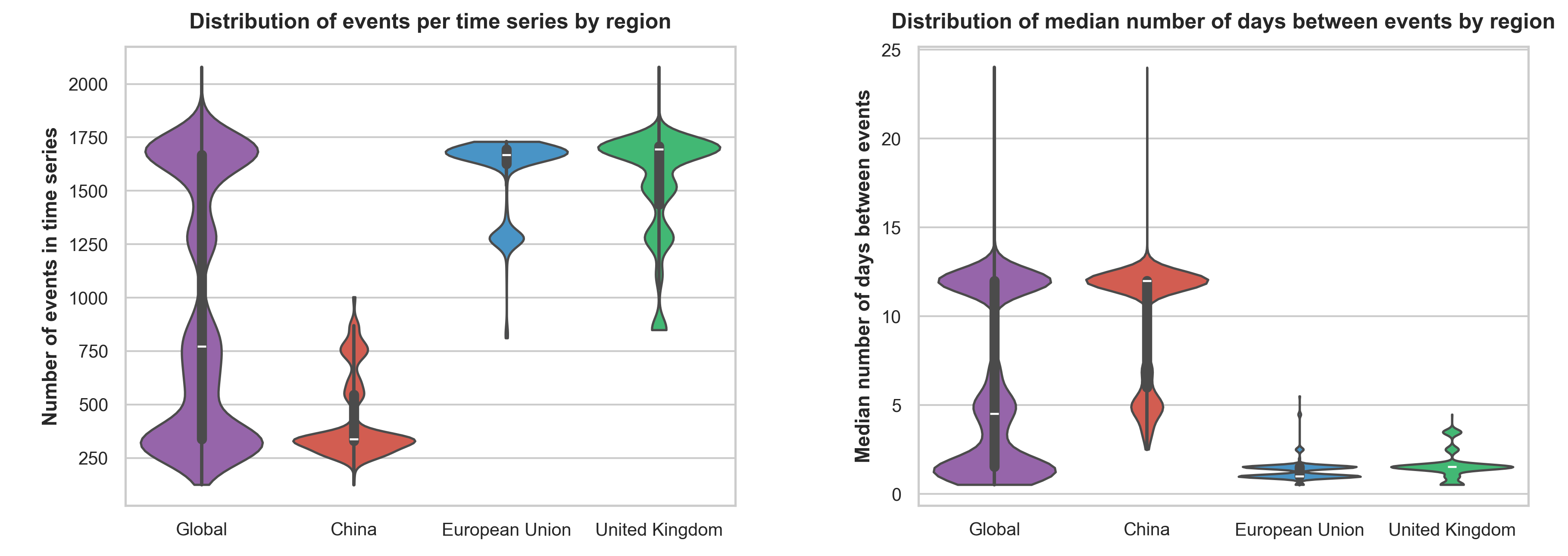}
	\caption{Global and regional distributions for the number of events per time series and the median number of days between time series events.}
	\label{fig:dataset_specs}
\end{figure}

The temporal resolution and thus number of events of each time series directly reflects Sentinel-1 acquisition frequency. During data set engineering, we focused on extracting the densest possible time series from the Sentinel-1 archive. As shown in Figure \ref{fig:dataset_specs}, temporal resolution varies geographically. Regions within the European Union show dense coverage corresponding to a higher number of events per time series due to higher acquisition frequencies and overlapping Sentinel-1 orbits, especially at higher latitudes (compare Figure \ref{fig:coverage}), resulting in revisit intervals of up to one day for certain locations. In contrast, regions such as the Chinese EEZ generally follow the mission’s default 12-day revisit interval of a single Sentinel-1 platform.

The provided 1D backscatter profiles are balancing feature richness, global completeness, and technical usability. The objective was to construct a globally consistent data set at maximal temporal resolution while preserving SAR features necessary for detailed deployment-phase analysis for a broad user audience, with different computational resources. The 14.8 million 1D backscatter profiles need about 5~GB of storage, enabling broad accessibility without requiring large-scale computational infrastructure. In contrast, storing the corresponding 2D Sentinel-1 image patches would increase data volume by an estimated factor of 25, significantly limiting usability to users with advanced computing resources.

\subsection{Global Offshore Wind Infrastructure Detection}\label{sec:res_owt_detect}
With the updated object detection workflow based on \citet{hoeser2022deepowt}, we identified 15,606 offshore wind infrastructure facilities across the entire global EEZ based on Sentinel-1 median composites. Of these, 15,100 were subsequently classified as offshore wind turbines in the time series analysis, see Figure \ref{fig:n_turbine_evolution}.

\begin{figure}
	\centering	\includegraphics[width=14cm]{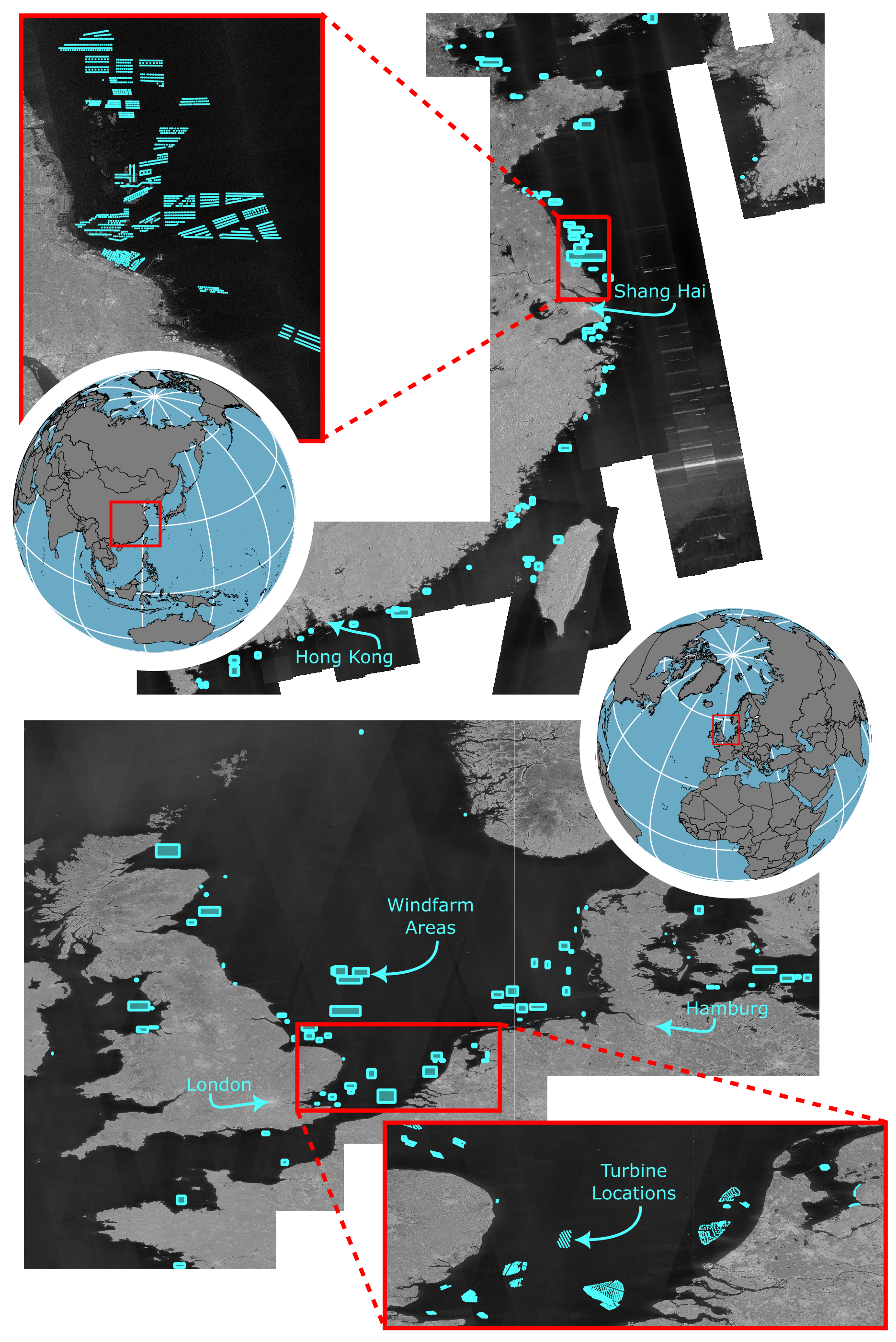}
	\caption{Results of the deep learning based offshore wind infrastructure detection workflow, showing bounding boxes for offshore wind farm areas proposed by a first CNN, and single offshore wind infrastructure locations within, provided by a second CNN. The upper part shows detections in Chinese waters, the lower part in the North Sea Basin, the two global wind energy hot spots.}
	\label{fig:detect_res_regions}
\end{figure}

At global scale, two dominant offshore wind hotspots are the Chinese EEZ and the North Sea Basin (NSB). Figure \ref{fig:detect_res_regions} illustrates the bounding boxes of offshore wind farm regions detected by the first-stage, which scans the global coastline and offshore areas. Zoomed examples further demonstrate detections of individual offshore wind infrastructure units identified by the second-stage within these candidate regions.

\begin{figure}
	\centering
	\includegraphics[width=\linewidth]{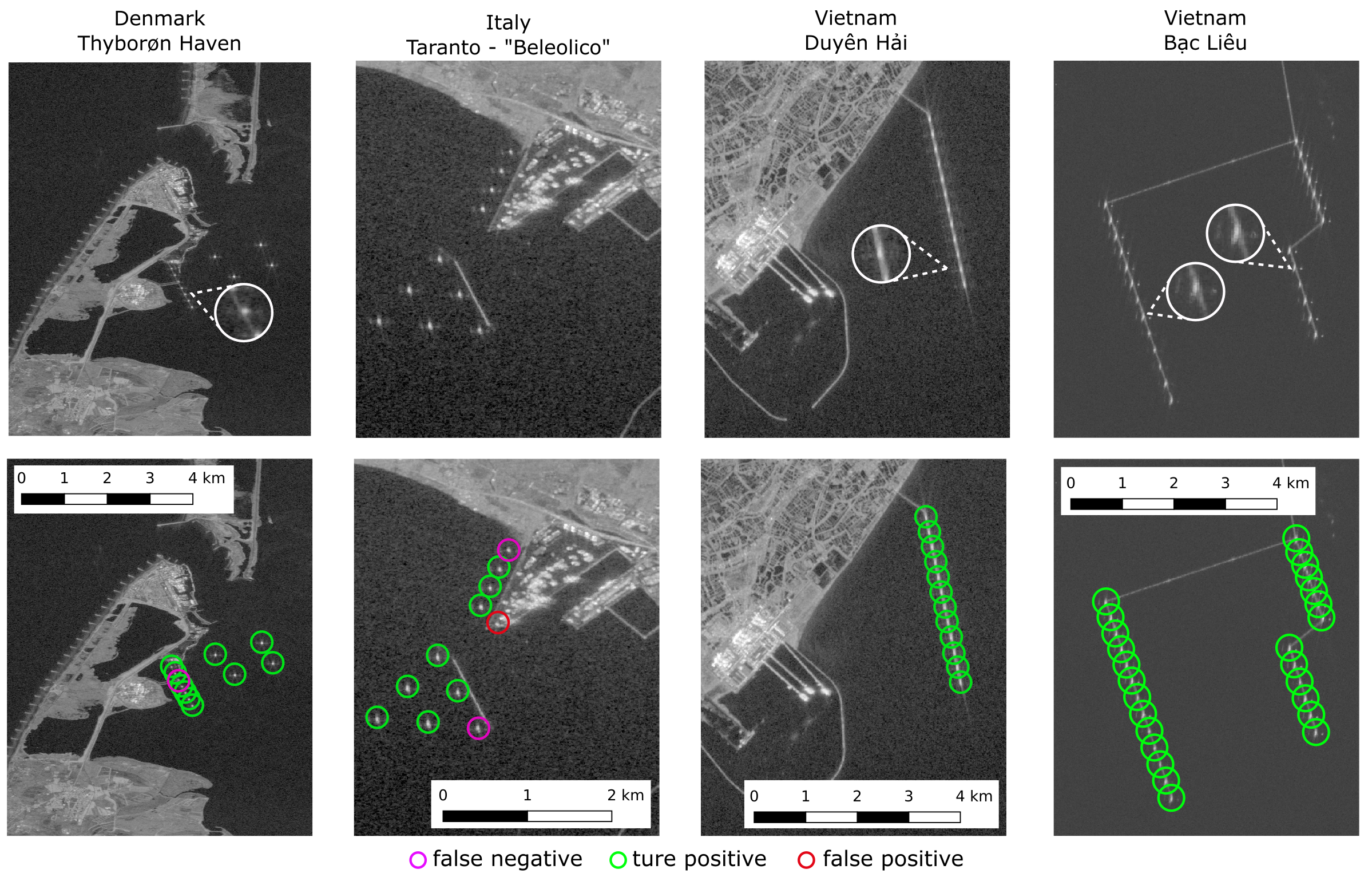}
	\caption{Detection results at challenging near coast and harbor environments.}
	\label{fig:detection_details}
\end{figure}

Ongoing expansion of the offshore wind energy sector introduce new spatial layouts that challenge detection models. Figure \ref{fig:detection_details} highlights complex environments such as harbors and near-coastal environments, where offshore wind turbines may be located close to piers or connecting infrastructure. Such sites pose recall challenges, increasing the risk of false omissions. Especially the wind farm layouts along the southeast coast of Vietnam represent challenging patterns that differ from typical patterns in the NSB or Chinese EEZ. At the same time, precision is challenged by coastal and harbor infrastructure with radar signatures similar to these layouts.

\begin{figure}
	\centering
	\includegraphics[width=\linewidth]{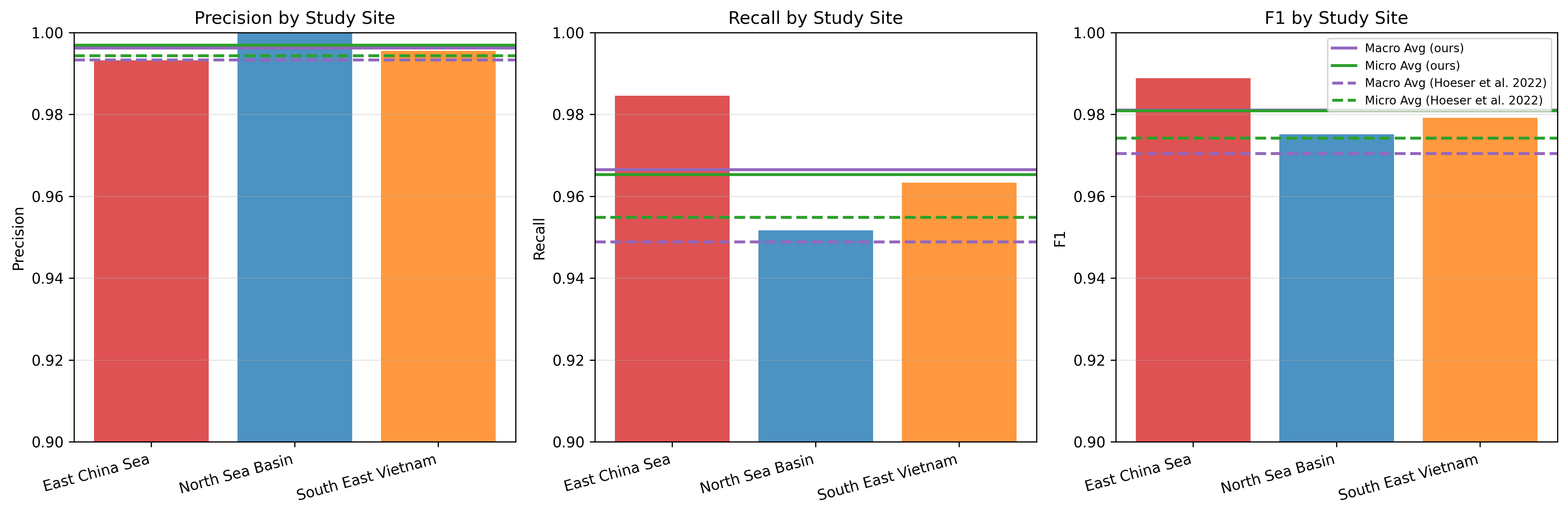}
	\caption{Detection performance of offshore wind energy infrastructure facilities at the three validation sites, the East China Sea, North Sea Basin and southeast Vietnam. Metrics from this study (\textit{ours}) are compared to detection performance of \citet{hoeser2022deepowt} where the same classes have been detected in Q2-2021.} 
	\label{fig:detection_performance}
\end{figure}

Detection performance across the three validation sites, the East China Sea, the NSB, and Southeast Vietnam is summarized in Figure \ref{fig:detection_performance} using Precision, Recall, and F1 score. Across all validation sites, performance ranges from 95.2\% to 100\% for individual metrics. The overall macro-averaged F1 score is 98.1\%. In comparison with the preceding study of \citet{hoeser2022deepowt} which is the only study also focusing on OWTs, other offshore infrastructure, and OWTs under construction as target classes, it outperforms the earlier results (97.04\% overall macro-averaged F1).

\begin{figure}
	\centering
	\includegraphics[width=\linewidth]{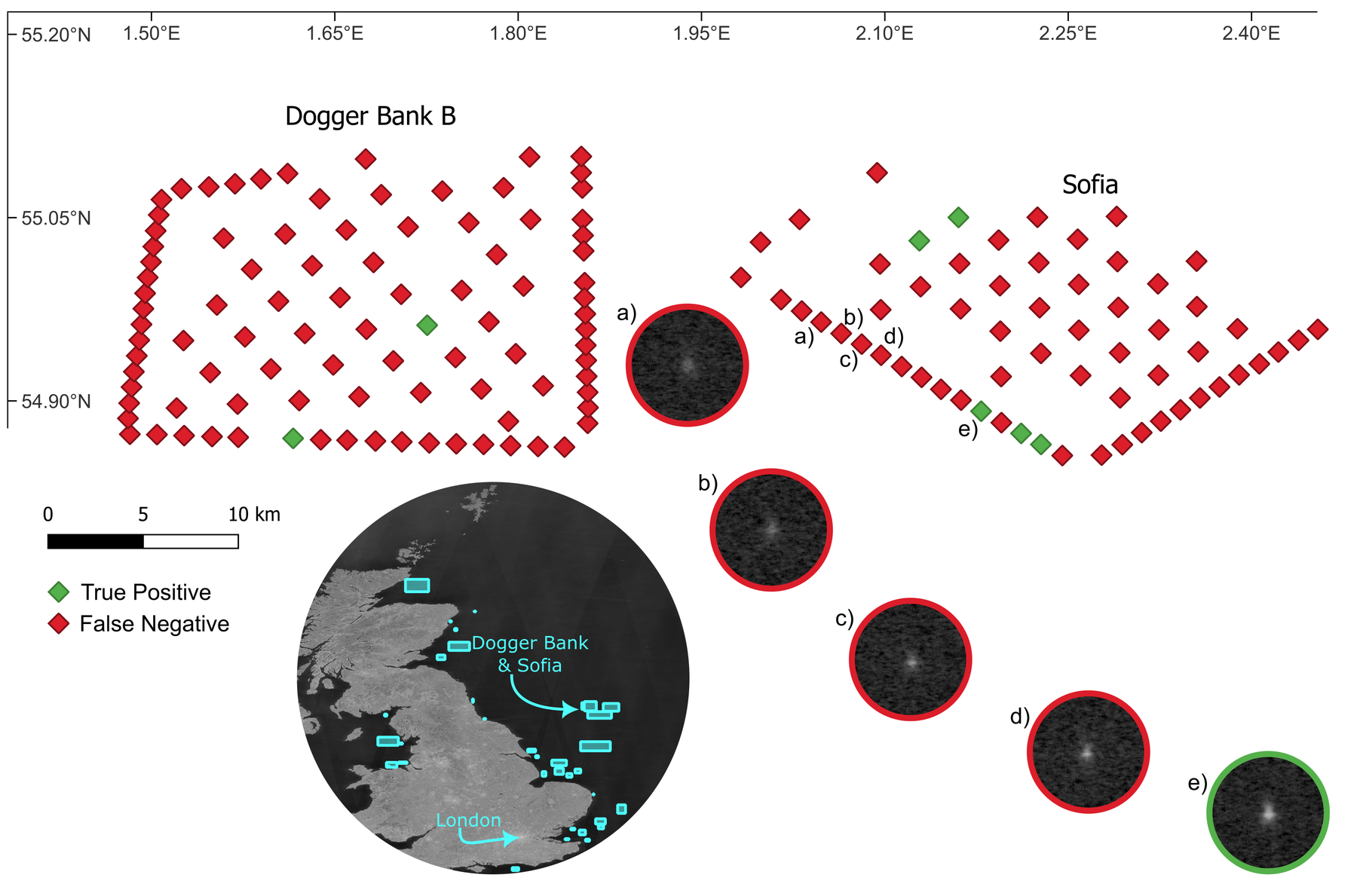}
	\caption{False omission cluster in the North Sea Basin, Dogger Bank, of offshore wind turbine foundations under construction during the median compositing time range in 2025Q1. False omissions can be attributed to less distinct radar backscatter patterns in the median composite, especially when a foundation's construction date is late in the compositing time range and is thus suppressed by pixels showing open water a)-d).}
	\label{fig:fn_report}
\end{figure}

The lowest metric observed in this study is a Recall of 95.2\% in the NSB. This is mainly caused by 262 false omissions (out of 5,419 ground-truth targets), most belonging to a large cluster of turbine foundations under construction at Dogger Bank. Figure \ref{fig:fn_report} shows this cluster and provides close insets of the missed turbine foundations and their appearance in the Sentinel-1 median composite used for detection. These insets demonstrate the challenging conditions of increasingly vague backscatter patterns, which indicate later foundation construction within the compositing time frame and a correspondingly decreasing visibility of the turbine foundation. See Section \ref{section:discussion} for further discussion of this issue.

\subsection{Time series classification}
For the 15,606 detected offshore wind infrastructure locations, we classified 15,606 corresponding time series with a total of 14,840,637 event labels. Six semantic classes are distinguished: water, turbine foundation, deployed turbine, support platform, mooring / construction vessel, and vessel.

Validation was performed on a benchmark subset of 553 expert-annotated time series containing 328,657 event labels. On the two important summary metrics, the area under the collapsed edit similarity-quality threshold curve (AUC) and the macro F1 score of event-based comparison, the proposed classifier performed with AUC of 0.785 and an F1 of 0.84.

\begin{figure}
	\centering
	\includegraphics[width=\linewidth]{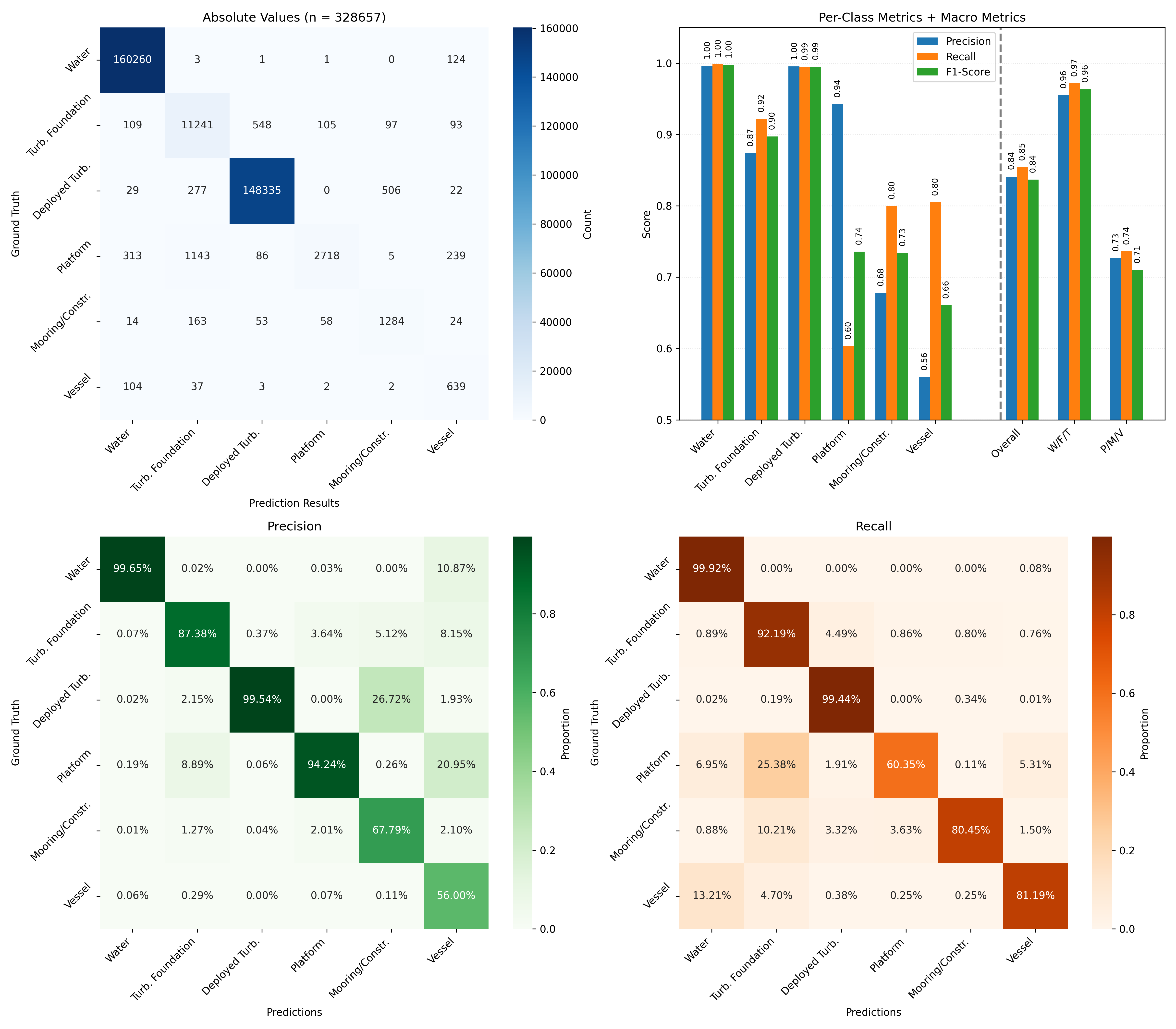}
	\caption{Event level, point-wise classification results as on predictions from the rule-based classifier. In the upper left metric over view of Precision, Recall and F1, Water, Turbine Foundation and Deployed Turbine are aggregated to W/F/T, Platform, Mooring/Construction vessel and Vessel are aggregated to P/M/V.}
	\label{fig:classification_metrics}
\end{figure}

Looking closer into the point wise evaluation of event classification, Figure \ref{fig:classification_metrics} reveals a clear separation in classification performance between deployment-related turbine classes (water, turbine foundation, deployed turbine; short W/F/T) and non-turbine or transient classes (support platform, mooring / construction vessel, vessel; short P/M/V). The turbine-related classes achieve a higher macro F1 score (0.96), whereas the platform and transient vessel-related classes show lower performance (macro F1 = 0.71) compared to the overall macro F1 score (0.84).

\begin{figure}
	\centering
	\includegraphics[width=12cm]{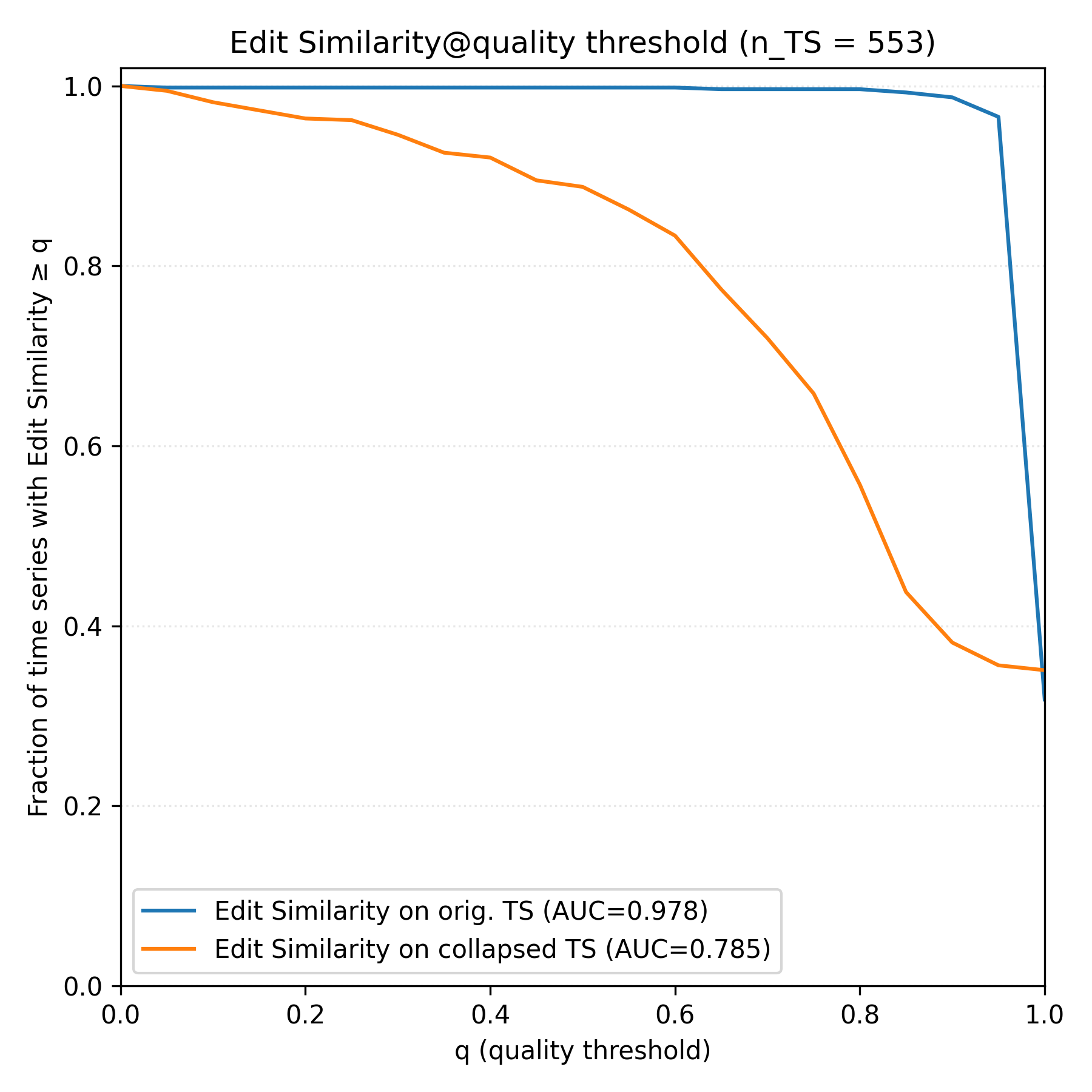}
	\caption{Fraction of Time Series with Edit Similarity scores higher than a given threshold at different threshold levels with corresponding AUC values on step-wise and collapsed classified time series.}
	\label{fig:edit_similarity}
\end{figure}

To complement single-event classification metrics, we evaluate sequential consistency using edit similarity. This metric quantifies how well predicted label sequences match the temporal structure of ground-truth sequences. Particularly informative is the collapsed edit similarity, see Figure \ref{fig:edit_similarity}. Its summary statistic AUC provides a measure of sequence and segment agreement and is therefore the most challenging and relevant benchmarking score for future studies. The rule-based classifier scores with an AUC of 0.785, with the steepest decline in successful edit similarity at 0.6 to 0.8 quality levels, stabilizing after the 0.9 quality level at 35\% of the classified time series. Meaning, 35\% of the classified time series have perfect sequential alignment with the ground truth data.

\begin{figure}
	\centering
	\includegraphics[width=\linewidth]{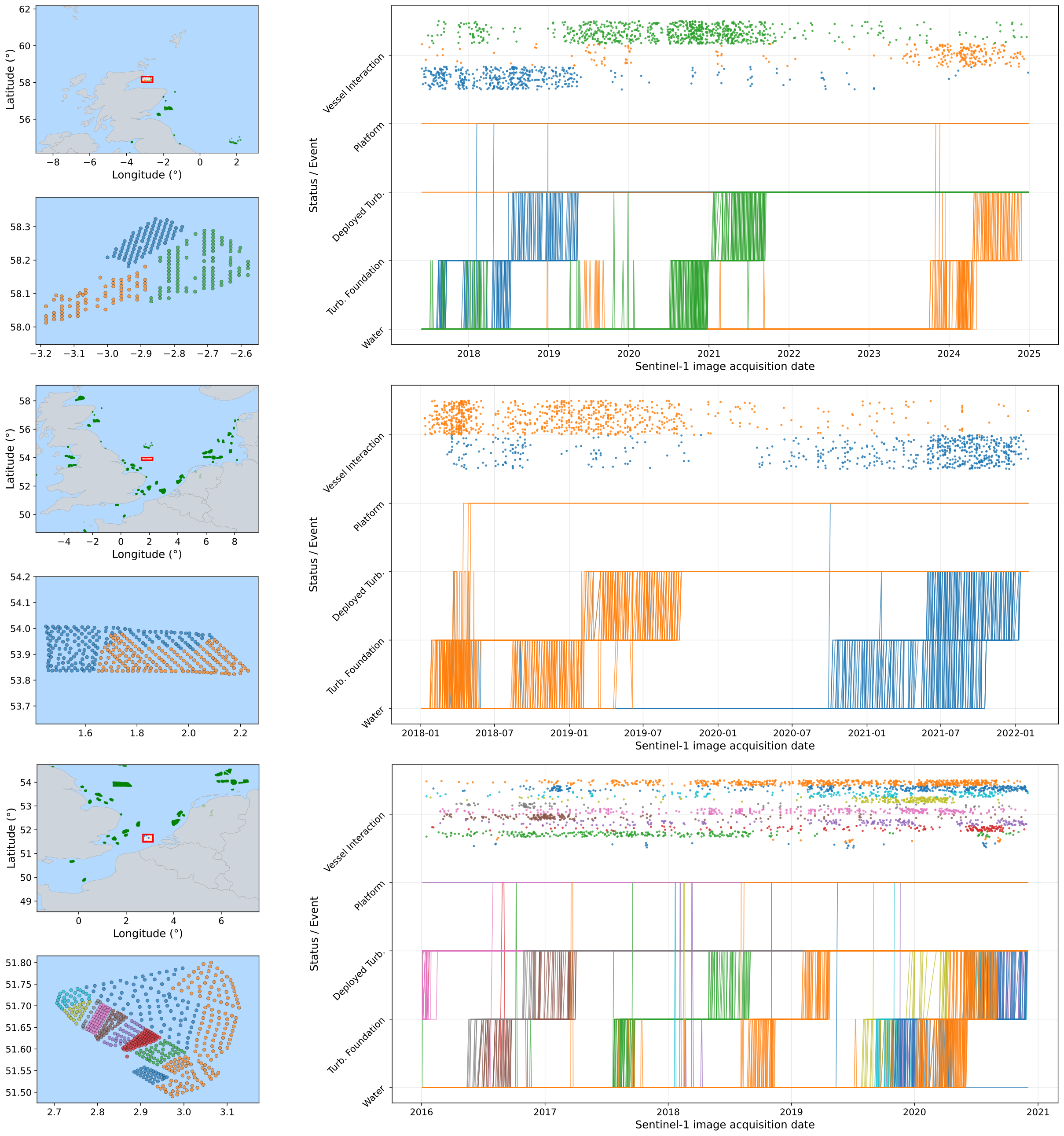}
	\caption{Example time series of three offshore wind farm clusters in the North Sea Basin. Each time series is a colored line corresponding to the location markers in the close-up overview map. Vessel related labels are markers in the time series with a random jitter for better visualization.}
	\label{fig:ts_results_EU}
\end{figure}

\begin{figure}
	\centering
	\includegraphics[width=\linewidth]{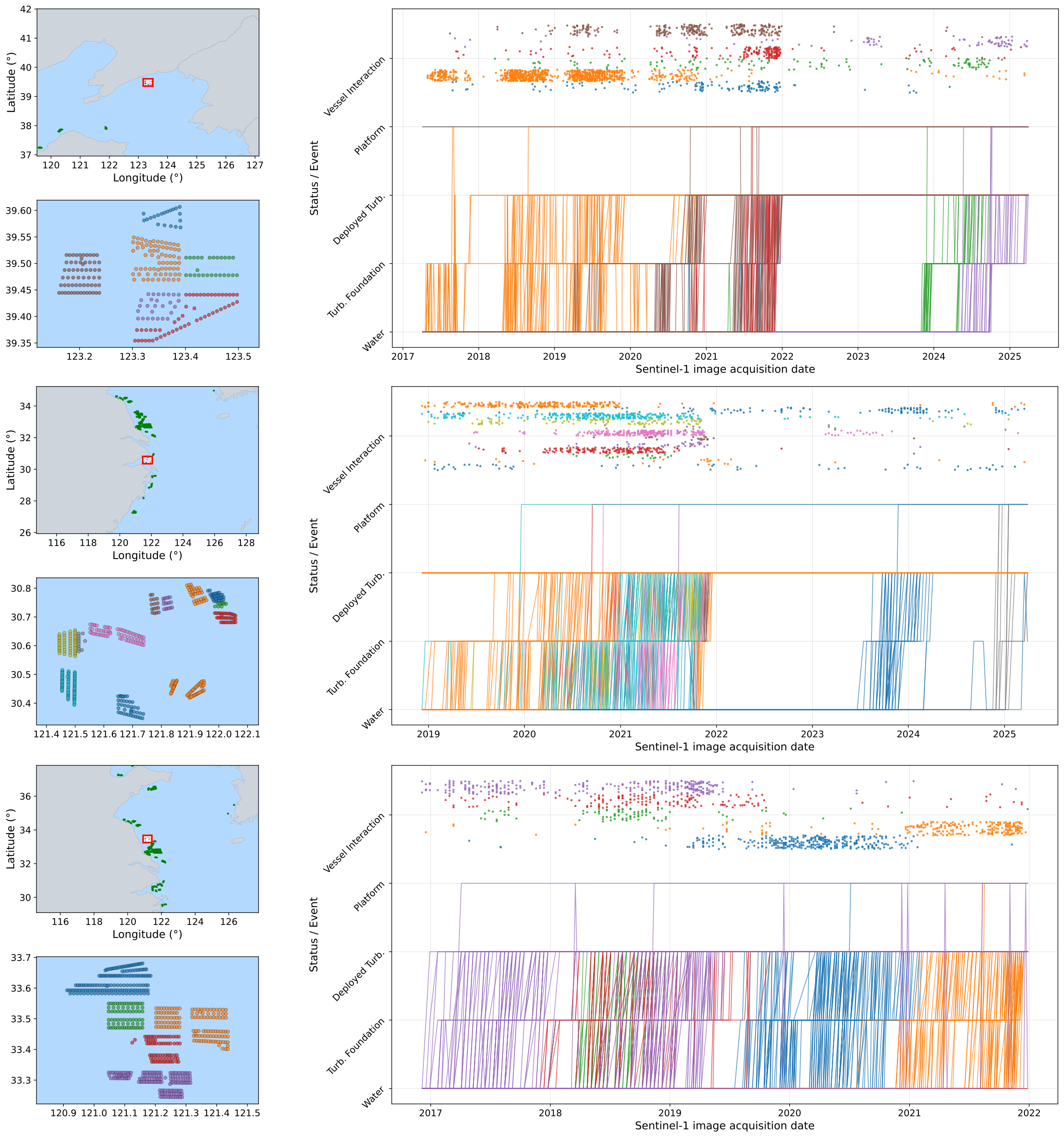}
	\caption{Example time series of three offshore wind farm clusters in the Chinese EEZ. Each time series is a colored line corresponding to the location markers in the close-up overview map. Vessel related labels are markers in the time series with a random jitter for better visualization.}
	\label{fig:ts_results_CN}
\end{figure}

In figures \ref{fig:ts_results_EU} and \ref{fig:ts_results_CN} we illustrate representative classified time series in spatial and temporal context. In the NSB, deployment phases are separated into distinct foundation construction and turbine installation stages, with a visible vessel interaction during peak construction periods. In contrast, examples from the Chinese EEZ show a less distinct separation between foundation and turbine deployment stages. The lag between foundation construction and turbine installation, commonly observed in NSB sites, is often absent, resulting in relatively shorter overall construction durations per turbine. However, vessel interaction patterns remain consistent across regions, with higher event density during construction compared to operational phases.

\begin{figure}
	\centering
	\includegraphics[width=\linewidth]{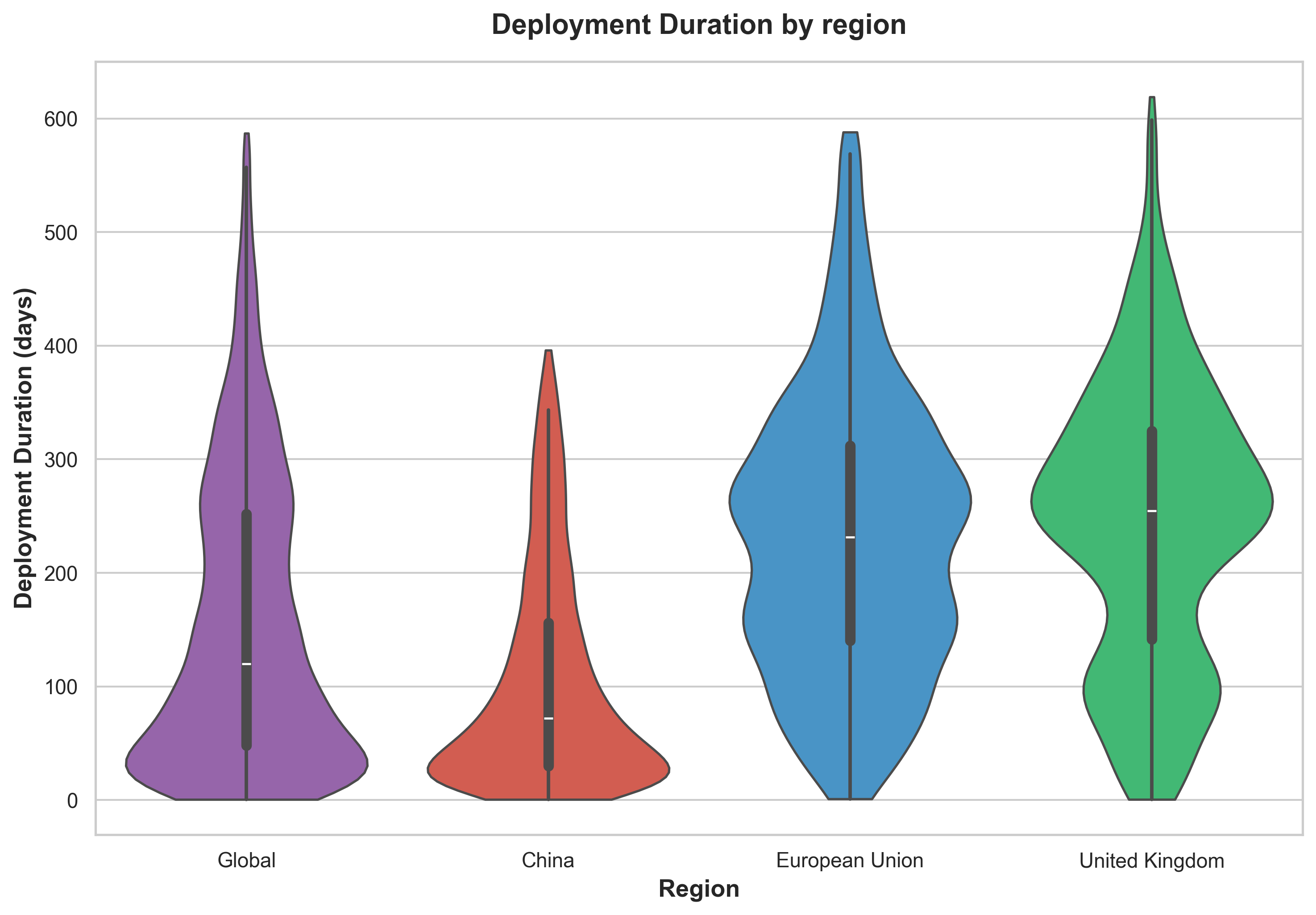}
	\caption{Distribution of offshore wind turbine deployment duration, derived as elapsed time measured from the first occurrence of a "turbine foundation" label to the first occurrence of a "deployed turbine" label, separated by regions.}
	\label{fig:deployment_duration}
\end{figure}

Using the classified time series, we derive regional deployment duration estimates by identifying the dates of first turbine foundation detection and first deployed turbine detection for each site and computing the elapsed time between these phases. The comparison of China, the European Union, and the United Kingdom shows significantly shorter median deployment durations within the Chinese EEZ (78 days) compared to the European Union (236 days) and the United Kingdom (258 days), see also Figure \ref{fig:deployment_duration}, which supports the observed differences in the example time series. These results demonstrate how the increased temporal resolution of the classified events enables large-scale analysis of regional deployment patterns.

\begin{figure}
	\centering
	\includegraphics[width=\linewidth]{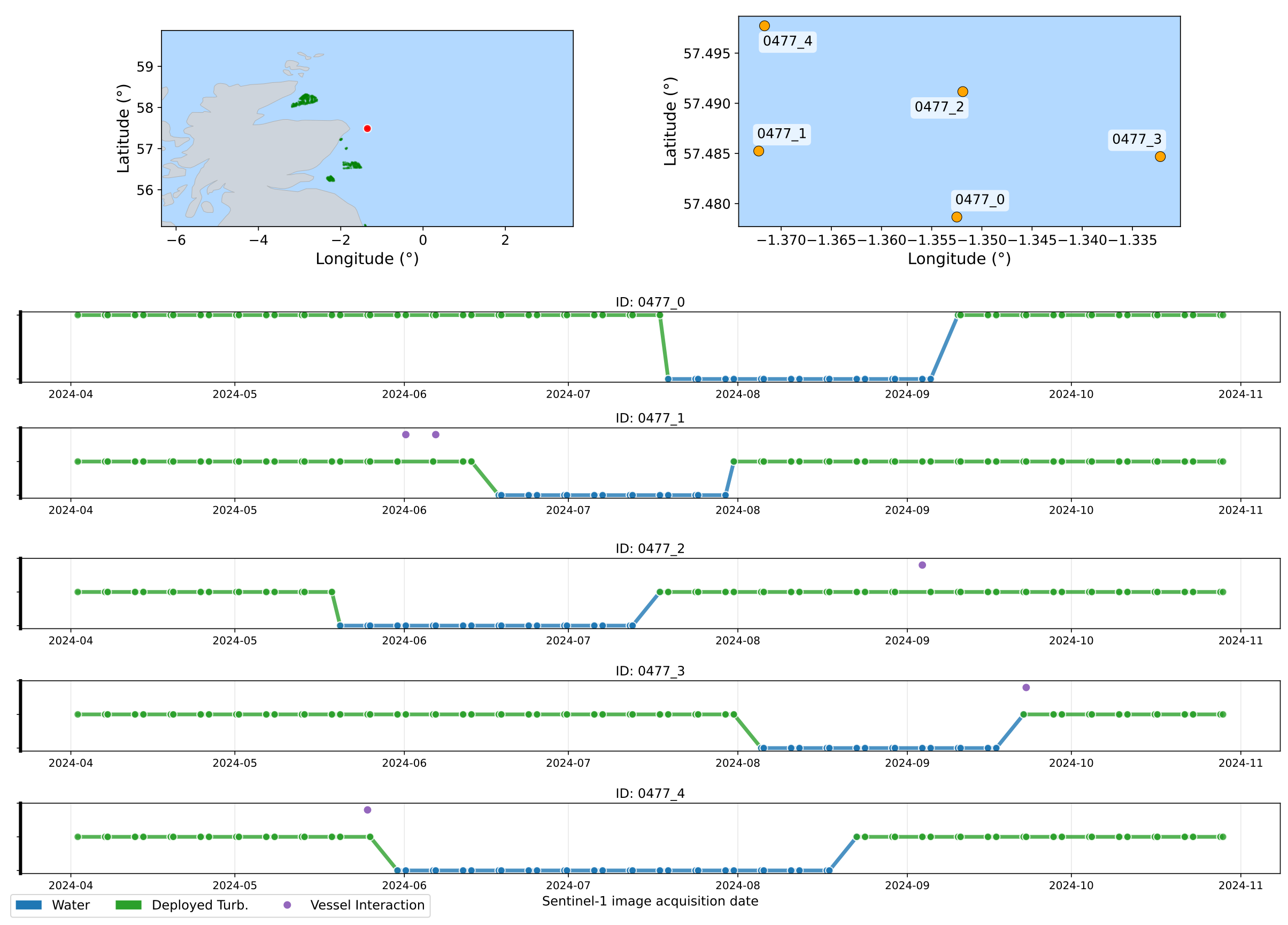}
	\caption{Visualization of the Hywind Scotland (HS) maintenance case study. HS is a floating turbine pilot project, in 2024 all 5 turbines have been towed away from their operational location due to planned maintenance. Reported start 2024-05-17 and end 2024-09-22 of the maintenance align well with classified time series patterns for all turbines.}
	\label{fig:hywind}
\end{figure}

To illustrate the sensitivity of the classified time series to operational events at a local scale, we present the case study of maintenance events at Hywind Scotland. Figure \ref{fig:hywind} presents results for the Hywind Scotland floating wind farm. In 2024, all five turbines were temporarily towed back to Norway for maintenance. According to official reports from Global Maritime \citep{GlobalMaritime2024_HywindMaintenance}, maintenance began on 2024-05-17, with the first turbine towed back to Norway reaching the port on 2024-05-24. The first turbine returned on 2024-07-05, and the final turbine on 2024-09-22.

These documented events align with the observed and classified time series in Figure \ref{fig:hywind}. The classification captures the absence of deployed turbine signatures during the maintenance periods. This case study provides an example, that the time series data of this publication provides temporally detailed insights exceeding those available from summarized reports, demonstrating the capability of the data set to detect and contextualize operational interventions at individual sites. Overall, the classified time series reveal deployment patterns, regional construction dynamics, and operational events, demonstrating the analytical potential of high-density SAR-based time series for monitoring offshore wind infrastructure.

\section{Discussion}
\label{section:discussion}
\subsection{Offshore wind infrastructure detection}
Over the past decade, Earth Observation based location mapping of offshore wind infrastructure has matured substantially, with methods which have been demonstrated for global-scale spatial mapping. The contribution of this work extents the field and lies in introducing high-density (1-12 days) temporal characterization of offshore wind infrastructure at global scale. In this sense, the focus shifts from asking "Where are turbines?" to asking "What is happening at each turbine location, and when?". By providing high density time series and baseline semantic labels, this study moves offshore wind infrastructure monitoring toward dynamic infrastructure intelligence rather than inventory mapping. However, the results also reveal that temporally dense monitoring remains fundamentally dependent on spatial detection. Bridging spatial localization and temporal analysis into a unified spatiotemporal framework represents a central challenge for upcoming research in offshore wind infrastructure monitoring from space.

The proposed 1D profile-based time series approach is highly effective once a location is known, but it requires initial spatial localization. CNN-based object detectors perform well in complex near-coastal and harbor environments, where heterogeneous background patterns demand high precision as reported by \citet{hoeser2022deepowt, Ding2024owtcn, Zhang2024gowtgeedeep, liu2024shandong, song2025seanson, LIU2026108706}. In contrast, CFAR- and rule-based approaches are often more sensitive to generic offshore metallic structures as demonstrated by \citep{e21060556, WONG2019111412, zhang2021gowt, Paolo2024} and may therefore be better suited to detecting early-stage turbine foundations. This distinction reflects a precision-recall trade-off, where CNN-based models tend to optimize precision in heterogeneous settings, and CFAR- or rule-based methods may increase recall for early construction signals but tend to be more prone to false alarms.

This trade-off becomes critical when aiming for early onset detection of turbine foundations. Early detection without prior location knowledge remains a conceptually unresolved problem. Future research must address how spatial detection methods can be designed to capture the very first construction signals without overwhelming downstream time series analysis or other methods to reduce false positives. Solutions such as hybrid CFAR-CNN cascades optimized specifically for early-stage foundations, or the integration of temporal anomaly detection directly at the spatial detection stage represent promising directions. Recent work, such as \citet{Paolo2024}, provides a starting point for such single-image detection frameworks, particularly when integrating SAR with AIS or optical data sources and cascading CFAR target detection and subsequent CNN-based classification. Since false omissions of turbine foundations are the bottleneck of the CNN detection process in this study, as reported in Section \ref{sec:res_owt_detect}, future improvements of the detection module will investigate the integration of such hybrid approaches. Additionally, the findings of this study make clear that aggregated Earth Observation imagery for detection introduces conceptual challenges when combined with temporal OWT event classification, that cannot be addressed by a new detection method alone. Instead, detection has to take place on single acquisitions or rolling aggregates to avoid missing the onset of infrastructure units when they are not yet represented in the aggregated data.

To advance towards an OWT detection process that aligns with the requirements of early-stage detection for dense temporal OWT event classification, a data corpus is urgently needed that provides not only derived detection results but also a benchmark data set enabling the thorough comparison of OWT detection studies. Currently, OWT detection is mostly reported by comparing derived products \citep{hoeser2022deepowt, Ding2024owtcn, LIU2026108706}, however, these products are often not temporally aligned, may focus on different spatial scales or sites, or address different problems in terms of output granularity, e.g., binary turbine detection versus the detection of foundation, deployed turbine, and support platform within a single model, see Table \ref{tab:review}. These differences render direct comparison ambiguous, especially given that, due to the maturity of the field, improvements occur within the top 5\% or even top 1\% range of typical detection scores. We argue that progress in the next phase of OWT detection would be substantially supported by the establishment of a benchmark data set with defined task categories for detection, similar as proposed in this study for time series based OWT event classification.

\subsection{Time series classification}

With respect to time series classification, the rule-based classifier presented here was intentionally designed as a relatable, training-data-free baseline. However, the implemented rule-based classifier has already reached considerable structural complexity, including multi-stage decision logic, sequential smoothing, segment-level corrections, and global label distribution refinements. This complexity suggests that the decision boundaries between infrastructure states are not trivial and may exceed what threshold-based logic can robustly encode. While the baseline achieves strong event-level performance, future improvements may require supervised deep learning.

The evaluation of time series classification highlights the performance gap between event-level accuracy and sequential consistency. Even when point-wise classification performance is strong, especially for turbine-related classes, the collapsed edit similarity score clearly shows that sequence-level agreement leaves room for improvement. This reflects the methodological limitations of hand-crafted rule-based classifiers, where sequential smoothing and refinements improves coherence but tends to risks over-regularization by enforcing deterministic transitions, or opposed to this, miss to refine overly volatile sequences to avoid over-regularization. Thus, future approaches may benefit from learning temporal transitions directly rather than relying on heuristic post-processing.

Taken together, this work demonstrates that spatial mapping of offshore wind infrastructure has reached a high degree of maturity, while fine grained temporal characterization is emerging as the new frontier. The presented data set and baseline classification model shift offshore wind infrastructure monitoring toward time series based event analysis. At the same time, spatial detection remains the bottleneck for early monitoring. The next generation of Earth Observation based offshore wind infrastructure monitoring will likely require detection approaches optimized for early onset signals and trained spatiotemporal models integrating detection and temporal event classification. Bridging spatial detection and temporal characterization into a unified spatiotemporal monitoring framework will be essential for achieving near-real-time, globally scalable monitoring of offshore wind infrastructure.

\section{Conclusion}

In this study we present an approach and data set for Earth Observation based offshore wind infrastructure event classification from high-temporal-resolution time series with a median temporal resolution ranging from 1 to 12 days depending on the location and available platforms at the given time. The data set contains open, analysis-ready, and globally distributed Sentinel-1 synthetic aperture radar (SAR) time series spanning 2016Q1–2025Q1, centered on 14,840,637 1D backscatter profiles compiled into 15,606 location-specific time series. These profiles provide a compact representation of global scale offshore wind infrastructure temporal dynamics, that preserves key SAR signatures while remaining accessible in storage and compute requirements.

For a spatial foundation, we update a global two-stage detection workflow and provide 15,606 offshore wind infrastructure locations for 2025Q1, with a macro F1 detection performance of 98.1\% over all target classes (Offshore wind turbines, offshore platforms, and offshore wind turbine foundations) and across three geographically diverse validation sites (North Sea Basin, China EEZ, and the South East Vietnamese Coast). To classify the 1D SAR backscatter profile time series at each of the detected locations, we introduce a rule-based classifier that assigns event-level semantic labels describing deployment and operational states, and we release a benchmark data set of 553 expert-annotated time series containing 328,657 event labels to support reproducible evaluation and method comparison. The baseline event classifier achieves a macro F1 score of 84\% over all event classes (water, turbine foundation, deployed turbine, support platform, mooring / construction vessel, vessel), and performs particularly well for turbine-related events (transition from water to turbine foundations, and transition from turbine foundations to deployed turbines, 96\% macro F1). Sequential similarity of the classified time series is measured with an area under the collapsed edit similarity score at quality thresholds curve with 0.785. Example analyses of the classified time series reveal differences in regional patterns concerning deployment duration with China showing significantly lower median deployment duration (78 days) compared to the European Union (236 days) and the United Kingdom (258 days). On a local scale we also show sensitivity to operational interventions such as maintenance-related towing events for floating offshore wind turbines captured by the classified time series events.

Beyond the data set contribution, the results highlight key directions for future work. Early-onset detection of wind infrastructure construction remains challenging, motivating research toward single-acquisition and hybrid detection approaches that balance recall for early detection of OWT foundations with precision in complex coastal environments. Concerning the introduced baseline model for time series event classification, the complexity required for a handcrafted rule-based classifier indicates that learned sequential models are promising future research areas.

Overall, the released data corpus establishes a foundation for future advancements in time series based offshore wind infrastructure monitoring from space by combining global coverage, maximum Sentinel-1 temporal resolution, analysis-ready time series, baseline predictions, and expert annotations as benchmark data set. It enables in-depth investigation of offshore wind deployment and operation at scale and supports the development, evaluation, and benchmarking of temporal analysis methods for independent monitoring of this rapidly expanding critical infrastructure.

\section*{Data Availability}
The data set is publicly available on Zenodo at 
\href{https://doi.org/10.5281/zenodo.5933966}{https://doi.org/10.5281/zenodo.5933966} 
\citep{hoeser_2026_18735421}.

\section*{Author contributions}
Conceptualization TH and CK;
Data curation TH;
Formal analysis TH;
Methodology TH;
Software TH;
Supervision CK and FB;
Validation TH;
Visualization TH;
Writing – original draft TH;
Writing – review and editing TH, FB, and CK.

\section*{Competing interests}
The contact author has declared that none of the authors has any competing interests.

\section*{Acknowledgements}
The authors gratefully acknowledge the Copernicus program of the European Space Agency (ESA) for providing free access to Sentinel-1 data. We also thank DLR’s Terrabyte team for their dedicated efforts in platform operations, which enabled the deep learning experiments, synthetic training data generation, and large-scale inference. We want to thank the two anonymous reviewers for their constructive and helpful comments.

\section*{Funding}
This research did not receive any specific grant from funding agencies in the public, commercial, or not-for-profit sectors.

\section*{Declaration of generative AI and AI-assisted technologies in the manuscript preparation process}
We acknowledge the use of Grammarly v1.2.124.1571, which includes a generative AI assistant for English language editing. All AI-generated text suggestions have undergone rigorous revision by the authors.

\newpage

\section*{Abbreviations}

\begin{tabular*}{\textwidth}{r@{\hspace{0.8em}}l@{\extracolsep{\fill}}}
AIS & Automatic Identification System \\
AUC & Area Under the Curve \\
CFAR & Constant False Alarm Rate \\
CNN & Convolutional Neural Network \\
CSV & Comma Separated Value \\
DLR & German Aerospace Center (Deutsches Zentrum für Luft- und Raumfahrt) \\
EEZ & Exclusive Economic Zone \\
ESA & European Space Agency \\
EO & Earth Observation \\
FN & False Negative \\
FP & False Positive \\
GDAL & Geospatial Data Abstraction Library \\
GEE & Google Earth Engine \\
GRD & Ground Range Detected \\
GTI & GDAL Tile Index \\
HPC & High Performance Computing \\
HS & Hywind Scotland \\
IW & Interferometric Wide \\
NRT & Near Real-Time \\
NSB & North Sea Basin \\
OWT & Offshore Wind Turbine \\
SAR & Synthetic Aperture Radar \\
TP & True Positive \\
TS & Time Series \\
VH & Vertical transmit, Horizontal receive \\
VV & Vertical transmit, Vertical receive \\
YOLO & You Only Look Once \\
\end{tabular*}

\bibliographystyle{abbrvnat}
\bibliography{references}

\end{document}